\documentclass[10pt,twocolumn,letterpaper]{article}

\usepackage{verbatim}
\usepackage[pagenumbers]{cvpr} 

\usepackage{multirow}
\usepackage{etoolbox}
\usepackage{listings}
\usepackage{xcolor}
\usepackage[table]{xcolor}
\usepackage[most,breakable]{tcolorbox}
\usepackage{makecell}
\usepackage[section]{placeins}
\usepackage{soul}

\lstdefinestyle{yaml}{
  basicstyle=\ttfamily\footnotesize,
  keywordstyle=\color{blue},
  stringstyle=\color{orange},
  commentstyle=\color{gray}\itshape,
  morekeywords={description, prompt, sub_agents, tools, input_proc, output_proc},
  backgroundcolor=\color{gray!10},  
  frame=single,  
  framerule=0.5mm, 
  rulecolor=\color{gray}, 
  breaklines=true, 
  xleftmargin=10pt, 
  xrightmargin=10pt 
}

\tcbuselibrary{listings}

\newtcolorbox{promptbox}{
  colback=gray!5,      
  colframe=gray!60,    
  boxrule=0.5pt,       
  arc=2pt,             
  left=6pt,right=6pt,top=4pt,bottom=4pt, 
  enhanced,
  breakable,
}

\usepackage{etoolbox}

\definecolor{cvprblue}{rgb}{0.21,0.49,0.74}
\usepackage[pagebackref,breaklinks,colorlinks,allcolors=cvprblue]{hyperref}
\usepackage{bbding}

\def\paperID{} 
\def\confName{CVPR}
\def\confYear{2026}

\title{MiroFlow: Towards High-Performance and Robust Open-Source Agent Framework for General Deep Research Tasks}

\author{Shiqian Su$^{1,2\dagger}$, Sen Xing$^{1,2\dagger}$, Xuan Dong$^{1,2\dagger}$, Muyan Zhong$^{1,2\dagger}$, Bin Wang$^{2}$, Xizhou Zhu$^{2}$, Yuntao Chen$^{2}$, \\Wenhai Wang$^{2}$, Yue Deng$^{2}$, Pengxiang Zhu$^{2,3}$, Ziyuan Liu$^{1,2}$, Tiantong Li$^{1,2}$, Jiaheng Yu$^{2}$, Zhe Chen$^{2,4}$, \\Lidong Bing$^{2}$, Jifeng Dai$^{1,2}$\textsuperscript{\Envelope}\\
$^1$Tsinghua University~~~
$^2$MiroMind AI~~~
$^3$National University of Singapore~~~
$^4$Nanjing University\\
}

\begin{document}
\maketitle
{
  \renewcommand{\thefootnote}{}
  \footnotetext{$\dagger$ Equal contribution.}
  \footnotetext{This work was done while Shiqian Su, Sen Xing, Xuan Dong, Muyan Zhong, Pengxiang Zhu, Ziyuan Liu, Tiantong Li and Zhe Chen were interns at MiroMind AI.}
  \footnotetext{Codes are available at \url{https://github.com/MiroMindAI/miroflow}}
}
\begin{abstract}

Despite the remarkable progress of large language models (LLMs), the capabilities of standalone LLMs have begun to plateau when tackling real-world, complex tasks that require interaction with external tools and dynamic environments.
Although recent agent frameworks aim to enhance model autonomy through tool integration and external interaction, they still suffer from naive workflows, unstable performance, limited support across diverse benchmarks and tasks, and heavy reliance on costly commercial APIs.
In this work, we propose a high-performance and robust open-source agent framework, termed MiroFlow, which incorporates an agent graph for flexible orchestration, an optional deep reasoning mode to enhance performance, and a robust workflow execution to ensure stable and reproducible performance.
Extensive experiments demonstrate that MiroFlow consistently achieves state-of-the-art performance across multiple agent benchmarks, including GAIA, BrowseComp-EN/ZH, HLE, xBench-DeepSearch, and notably FutureX. We hope it could serve as an easily accessible, reproducible, and comparable baseline for the deep research community.

\end{abstract}    

\section{Introduction}
\label{sec:intro}

\begin{figure*}[h]
\centering
\includegraphics[width=1.\linewidth]{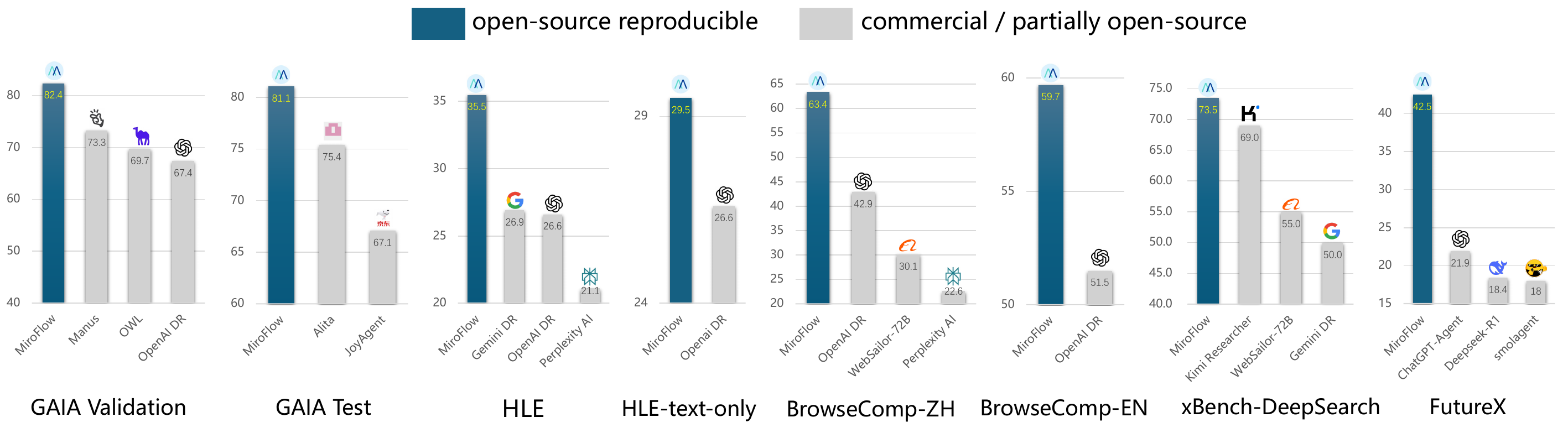}
\caption{
\textbf{Overall performance of MiroFlow on representative deep research benchmarks.} MiroFlow, a \textit{high-performance} and \textit{robust open-source} agent framework, achieves \textit{reproducible state-of-the-art} results across all benchmarks, consistently outperforming existing open-source and commercial agent systems. All MiroFlow results are obtained with a single unified configuration without any task-specific tuning, demonstrating strong generality and adaptability across heterogeneous deep research tasks. 
}
\label{fig:performance}
\end{figure*}

The rapid advancement of Large Language Models (LLMs) has dramatically expanded the frontiers of artificial intelligence.
By leveraging massive datasets~\cite{cc, c4, refineweb} and scaling parameters to unprecedented levels, LLMs~\cite{gpt3,llama,qwen,deepseek} have shown remarkable emergent abilities across diverse tasks.
However, real-world applications are far more complex than short-term question answering or closed-form inference.
Most mainstream LLMs still operate in a self-contained manner, without interaction with external tools or environments, and their performance has begun to plateau, making them inadequate for complex real-world scenarios that require multi-step planning, dynamic reasoning, and tool coordination.

Recent open-source agent frameworks~\cite{smolagents,hu2025owl,yu2025aworldorchestratingtrainingrecipe,tongyideepresearchteam2025tongyideepresearchtechnicalreport,zhang2025agentorchestraorchestratinghierarchicalmultiagent,liu2025joyagentjdgenietechnicalreportgaia} have begun to bridge this gap by equipping LLMs with planning capabilities and external tool use. However, three persistent challenges continue to limit their reliability and adoption in complex deep research scenarios. (1) \emph{Inflexibility.} Most systems are designed for single-purpose workflows with hard-coded pipelines, limiting their generalization to heterogeneous tasks; (2) \emph{Instability.} Due to long reasoning chains, stochastic sampling, tool invocation failures, and search result randomness, model performance fluctuates significantly and experimental results are often difficult to reproduce; This attribute also necessitates that the agent can comprehensively understand and utilize diverse tools, and handle various exceptional or uncertain situations. Only in this manner can reproducibility and fair comparison in research be ensured. (3) \emph{High Cost.} Integration with commercial APIs leads to substantial financial overheads, which hinder open research and limit the accessibility and scalability of agent-based systems.
Table~\ref{tab:gaia_scores_filled} lists several agent frameworks for comparison.

An ideal agent framework should therefore transcend rigid, hard-coded pipelines and instead empower users to flexibly configure the agent’s operational workflow, supporting the dynamic composition of reasoning modules and external tools tailored to specific task requirements. 
Furthermore, robustness is a fundamental design principle. 
A well-designed mechanism should be in place to ensure systematic fallback and verification, guaranteeing reliable execution under uncertainty.
On this foundation, transparency is also crucial. By integrating open-source tools or establishing transparent evaluation protocols, such a framework can significantly lower research costs, promote accessibility, and accelerate community-driven innovation in agent-based systems.

In this work, we present MiroFlow, a high-accuracy and robust open-source agent framework that achieves reproducible state-of-the-art performance across diverse deep research tasks.
Building upon the principles outlined above, MiroFlow introduces several key innovations:
First, we propose a hierarchical agent framework driven by an agent graph, which enables users to flexibly compose and configure agents and their components in a top-down manner, adapting the workflow to specific task requirements.
Second, an optional heavy-reasoning mode allows the agent to perform deeper reasoning and self-verification in high-stakes or complex scenarios, thereby enhancing overall reasoning depth and performance.
Third, a robust workflow mechanism systematically mitigates stochastic errors and stabilizes long reasoning chains, ensuring consistent and reproducible performance across repeated executions.
Finally, MiroFlow supports a range of open-source tools, libraries and benchmarks, enabling cost-efficient, transparent, and scalable deployment in diverse deep research environments.

We conduct extensive experiments to evaluate the effectiveness of MiroFlow. As shown in Figure \ref{fig:performance}, MiroFlow achieves state-of-the-art performance across multiple representative agent benchmarks. In addition, we see that the same codebase can be seamlessly applied across different benchmarks without any task-specific tuning, highlighting the generality, adaptability, and reliability of the proposed framework. These findings confirm that MiroFlow serves as a unified, transparent, and reproducible foundation for advancing future research in large-scale, agent-based systems.

In summary, our contributions are as follows:

(1) \textbf{We present MiroFlow}, a high-performance and robust open-source agent framework. MiroFlow adopts a hierarchical architecture (\emph{i.e.}, control, agent, and foundation tiers), where the control tier coordinates the interaction between the agent and foundation tiers, enabling general task scheduling and modular extensibility.

(2) \textbf{To enhance the framework’s flexibility, stability, and accuracy}, we introduce three synergistic components: the agent graph, a robust workflow, and a heavy-reasoning mode.
The agent graph provides a flexible interface for constructing task-specific execution graphs. The robust workflow incorporates task normalization, retry mechanism and fault isolation to mitigate stochastic fluctuations in multi-step reasoning, while the heavy-reasoning mode further improves consistency and precision for complex problems.

(3) 
\textbf{We evaluate MiroFlow on multiple representative agent benchmarks}, including GAIA~\cite{MialonF0LS24gaia}, BrowseComp-EN/ZH~\cite{wei2025browsecompsimplechallengingbenchmark,zhou2025browsecompzhbenchmarkingwebbrowsing}, HLE~\cite{phan2025humanitysexam}, xBench-DS~\cite{chen2025xbenchtrackingagentsproductivity}, and notably FutureX~\cite{zeng2025futurexadvancedlivebenchmark}.
The results demonstrate that MiroFlow achieves reproducible and consistent state-of-the-art performance across diverse tasks, 
highlighting its superior generality and reliability. We also conduct extensive ablation to provide insights for designing future agent systems and offer a comprehensive empirical foundation for the community.

We believe that MiroFlow lays a solid foundation for agent systems in deep research, effectively bridging the gap between open-source large language models and commercial LLM systems such as GPT-5.

\section{Related Work}
\label{sec:related_work}

\subsection{Large Language Model}

Large language models (LLMs) have rapidly scaled, driven by the empirical neural scaling law~\cite{kaplan2020scaling}. Built on the Transformer architecture~\cite{vaswani2017attention} and pre-trained on massive corpora~\cite{cc, c4, refineweb}, these models show predictable improvements as parameters and data grow, exemplified by the evolution of the GPT series~\cite{gpt1,gpt2,gpt3,gpt4,gpt5}. The emergence of open-source LLMs like LLaMA~\cite{llama}, Mistral~\cite{mistral7b}, Qwen~\cite{qwen}, DeepSeek~\cite{deepseek}, and GLM family~\cite{glm130b} has democratized access to cutting-edge AI, driving both academic and industrial growth.

Recent research has shifted focus from scaling parameters to enhancing \emph{reasoning} for complex tasks, such as DeepSeek-R1~\cite{deepseek_r1_2025}, OpenAI o3/o4-mini~\cite{openai_o3_o4mini_2025}, and Claude 3.7 Sonnet (Thinking)~\cite{anthropic_claude37_system_card_2025}, QwQ-32B~\cite{qwq32b}, GLM-4.5~\cite{glm4-5}, Kimi K1.5~\cite{kimi1-5}, which integrate reinforcement learning and inference-time scaling to improve reasoning accuracy. These advances, which combine chain-of-thought reasoning~\cite{wei2022chain}, self-reflection, and adaptive thinking, have led to significant improvements in cross-domain benchmarks~\cite{aime2025,cobbe2021gsm8k,math500_2023,hendrycks2020mmlu,lightman2023letsverify}. Alongside reasoning improvements, there has been a shift towards multimodal LLMs (MLLMs) which process text, images, and videos. MLLMs like LLaVA~\cite{li2024llavaonevisioneasyvisualtask}, QwenVL~\cite{qwen}, InternVL~\cite{internvl3-5} and MiMo-VL~\cite{mimo-vl} excel at cross-modal reasoning, enabling tasks such as visual grounding~\cite{plummer2015_flickr30kentities} and VQA~\cite{singh2019_textvqa}.

Although some progress has been made in recent years, ``self-contained" large language models (LLMs) are encountering diminishing returns, particularly in short-horizon QA tasks. These models are constrained by outdated knowledge and lack external interaction through tools or environments, leading to inherent limitations when dealing with complex problems. Real-world tasks are often complex and dynamic, and relying solely on ``self-contained" capabilities is no longer sufficient. As a result, the performance of these models is approaching saturation, especially in short-term QA. 
To address more complex real-world tasks, future LLMs must function as systems rather than standalone models. They will require stronger interactive capabilities, enabling adaptive reasoning through tool use, feedback mechanisms, and multi-agent collaboration.

\subsection{LLM-Based Agents}

To overcome the limitations of self-contained language models, recent research has shifted toward LLM-based agents, which reframe the model as an autonomous system capable of perception, reasoning, decision-making, and action~\cite{luo2025largelanguagemodelagent}.
These agents typically consist of two components: an agent foundation model, which provides core capabilities like reasoning, planning, and tool use, and an agent framework, which handles task decomposition, state management, and external feedback.
Together, these components transform an LLM from a passive knowledge source into an interactive problem solver.

Recent advances in foundation models focus on enhancing agentic capabilities, particularly through inference-time reasoning and native tool-use. Models now generate multi-step thinking traces for more reliable long-term reasoning and planning~\citep{deepseek_r1_2025, openai_o3_o4mini_2025, comanici2025gemini25pushingfrontier, anthropic_claude37_system_card_2025}. Additionally, tool-use is a core feature of modern agents, with function calling supported across proprietary and open-source models~\citep{openai2024gpt4technicalreport, openai_o3_o4mini_2025, comanici2025gemini25pushingfrontier, anthropic_claude37_system_card_2025, qwen3technicalreport, grattafiori2024llama3herdmodels}. Recent models like DeepSeek V3.1~\cite{deepseekai2024deepseekv3technicalreport}, Qwen3~\cite{qwen3technicalreport}, tongyi~\cite{tongyideepresearchteam2025tongyideepresearchtechnicalreport}, and Kimi K2~\cite{kimiteam2025kimik2openagentic} use multi-stage post-training pipelines to enhance tool-augmented reasoning and decision-making.

In parallel, a wide range of agent frameworks have been developed to orchestrate foundation models. Early systems primarily focused on augmenting LLMs with external tool access.
Toolformer~\cite{toolformer} fine-tuned a model to self-supervise its own API calls, whereas Visual ChatGPT~\cite{wu2023visualchatgpttalkingdrawing} and HuggingGPT~\cite{shen2023hugginggpt} treated the LLM as a central controller that delegates tasks to specialist models.
The introduction of the ``Thought–Action–Observation'' loop by ReAct~\cite{yao2023react} marked a shift toward more general reasoning agents, later popularized by AutoGPT~\cite{yang2023autogptonlinedecisionmaking}.
More recent efforts focus on multi-agent architectures and end-to-end agents that handle planning, reasoning, action, and multimodal understanding. For example, Deep Research systems~\cite{OpenAI2025DeepResearch, Google2024GeminiDeepResearch, Perplexity2025DeepResearch} and open-source platforms like OWL~\cite{hu2025owl}, OpenHands~\cite{wang2025openhands}, and AWorld~\cite{yu2025aworldorchestratingtrainingrecipe} aim to provide extensible frameworks for planning, tool use, and autonomous reasoning. Despite these advancements, LLM-based agents still face significant challenges, including rigid domain-specific workflows, unstable behavior with inconsistent outcomes, and reliance on commercial models and APIs leads to high operational expenses.

\section{MiroFlow Agent Framework}

\subsection{Overall Architecture}
\label{sec:overall-arch}

\begin{figure*}[htb]
\centering
\includegraphics[width=0.7\textwidth]{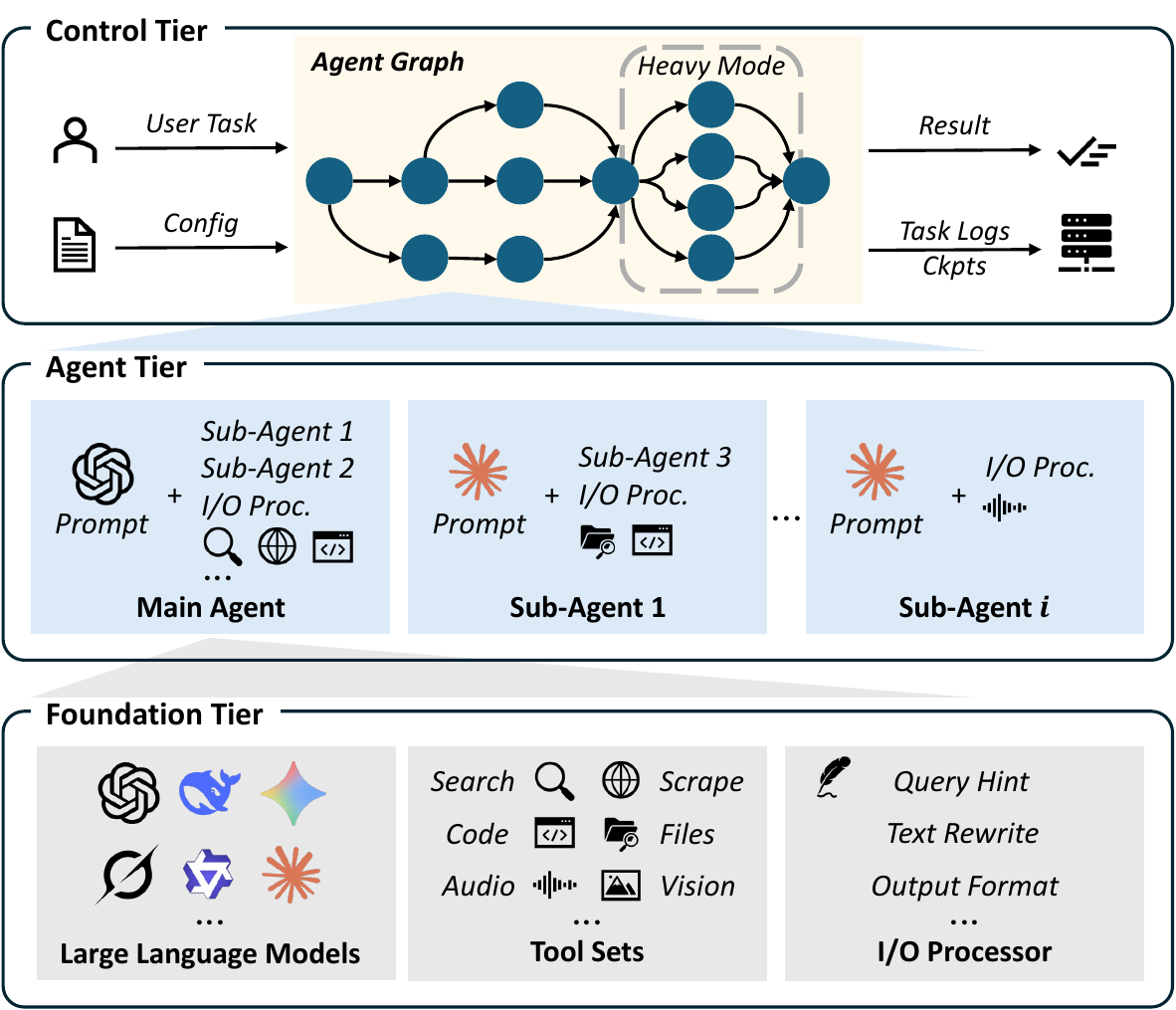}
\caption{
\textbf{Overview of the three-tier hierarchical MiroFlow framework architecture.} \textbf{Foundation tier} provides reusable core components, including LLM backends, MCP-based tool sets, and generic input–output processors, which supply the basic capabilities required by all agents. \textbf{Agent tier} defines a set of agent nodes constructed by combining foundation-tier components with specific prompts. Each node is provided with a list of agents it can call. These nodes communicate through structured messages and can be flexibly instantiated or extended. \textbf{Control tier} assembles multiple agent nodes into an agent graph and orchestrates the end-to-end workflow: user queries enter the graph and are processed through coordinated agent interactions and tool calls, while the controller maintains task logs and checkpoints for reproducibility, supports a heavy-reasoning mode to improve accuracy, and incorporates workflow-level robustness enhancements to ensure smooth and predictable execution.
}
\label{fig:overall}
\end{figure*}

MiroFlow is a high-precision, robust, and performance-reproducible open-source research agent framework designed to extend the capabilities of large language models (LLMs) by transforming them into a collaborative system of agents and tools.
As shown in Figure~\ref{fig:overall}, MiroFlow adopts a three-tier hierarchical architecture, consisting of the control tier, agent tier, and foundational tier, which are responsible for the orchestration of workflows, the behavioral logic of agents, and the foundational components supporting the agents, respectively.

\noindent\textbf{Control Tier.} As the controller of the system, this tier is responsible for orchestrating the overall flow of tasks based on user input and predefined agent graph configurations. It manages coordination between agents. The controller supervises the entire process, ensuring the correct sequence of operations, maintaining logs and checkpoints for reproducibility, and ensuring the smooth and predictable operation of the entire pipeline. Based on the agent graph, a heavy-reasoning mode is introduced to enhance accuracy by increasing inference time. In addition, enhancements to the robustness of the overall workflow ensure greater system reliability and reproducibility.

\noindent\textbf{Agent Tier.} The agent tier is the core of MiroFlow. It consists of multiple agent nodes, each representing an independent work unit. Each agent node contains its own context, prompts, base LLM, toolset, and input-output processors, and communicates with other accessible agent nodes through structured messages to collaborate on task completion. The agent tier is separate from the control tier, eliminating agent dependencies and certain supporting modules. This design provides scalability, allowing users to easily define, manage, and adjust agents as needed.

\noindent\textbf{Foundation Tier.} The foundation tier supports the MiroFlow framework by providing the essential core components and infrastructure required by the agent tier. It offers support for various backend models used by agent nodes, including popular LLMs such as GPT~\cite{gpt5}, Claude~\cite{anthropic_claude37_system_card_2025} and Qwen~\cite{qwen3technicalreport}, along with the necessary toolsets, basic input-output processing functions and resources for task execution. The primary responsibility of the foundation tier is to ensure the stability, scalability, and efficiency of both the backend models and tools, while also delivering seamless access interfaces for the agent tier.

\noindent\textbf{Example Pipeline.} Taking deep research as an example, a main agent node serves as the entry point for the entire workflow and is responsible for task decomposition, subtask delegation, and result synthesis. It coordinates with other agents equipped with search, web-reading, or coding capabilities. This collaborative structure yields the following pipeline:
(1) \emph{Query Augmentation.} During the main agent’s input processing, the user’s query is first analyzed by an LLM to identify intent and enrich the request, enabling a more accurate understanding of the requirements;
(2) \emph{Task Planning.} The main agent formulates a detailed execution plan based on the enhanced query content, coordinating the entire workflow, which includes invoking different tools, assigning tasks to sub-agents, and driving task progress;
(3) \emph{Agent Delegation.} For complex or specialized tasks, the main agent delegates subtasks to agents with relevant expertise (\eg, a browsing agent). These agents can independently plan and execute their tasks, invoke necessary tools, and, if needed, further delegate work to other agents.
(4) \emph{Tool Calling.} When external functionalities need to be invoked, agents connect to the MCP~\cite{mcp} server to obtain and use the corresponding specialized tools;
(5) \emph{Result Synthesis.} After task completion, the main agent synthesizes results from multiple information sources and performs output processing to ensure the final response is high-quality and aligned with the user’s requirements or specified formats.

\subsection{Agent Node}
\label{sec:agent-node}

Agent nodes are the fundamental execution units responsible for performing reasoning and tool interactions for tasks or subtasks. An agent node consists of the following key attributes:
(1) \emph{Description.} Provides a brief overview of the node's role and capabilities, guiding agent orchestration;
(2) \emph{Prompt.} System instructions tailored to specific roles, controlling reasoning behavior and output format;
(3) \emph{Sub-Agents.} Optional child agent nodes that this node may call to handle specialized subtasks when needed;
(4) \emph{Tools.} External APIs or functions registered through MCP, with each tool exposing formal input/output contracts;
(5) \emph{Input and Output Processors.} Includes the input processor and output processor, responsible for preprocessing the input and the output, ensuring input validation, normalization, and format consistency.

\subsection{Agent Graph}
\label{sec:agent-graph}

The Agent Graph defines how multiple agent nodes collaborate to complete a task. Unlike traditional chain-based or tree-based workflows, MiroFlow utilizes a directed graph structure, making the workflow more flexible and efficient. In a graph-based workflow, tasks can be executed in parallel or interwoven in various ways, without needing to follow a fixed sequence. This structure allows the framework to better accommodate complex task requirements.

MiroFlow employs a ``declare then define'' approach to describe workflows. In this approach, the main agent serves as the entry node, forming a topological structure responsible for initiating tasks and calling other agents and tools. Each agent can also define its own sub-agents and tools, creating a hierarchical topological structure. Through this topological structure, we can flexibly define dependencies between nodes, specifying which tasks must be executed sequentially and which can be executed in parallel. The entire system's workflow is described through the connections and dependencies between these nodes, ensuring that the execution process is both clear and highly scalable.

\subsection{Heavy-Reasoning Mode}
\label{sec:heavy-mode}

\begin{figure}[h]
\centering
\includegraphics[width=0.35\textwidth]{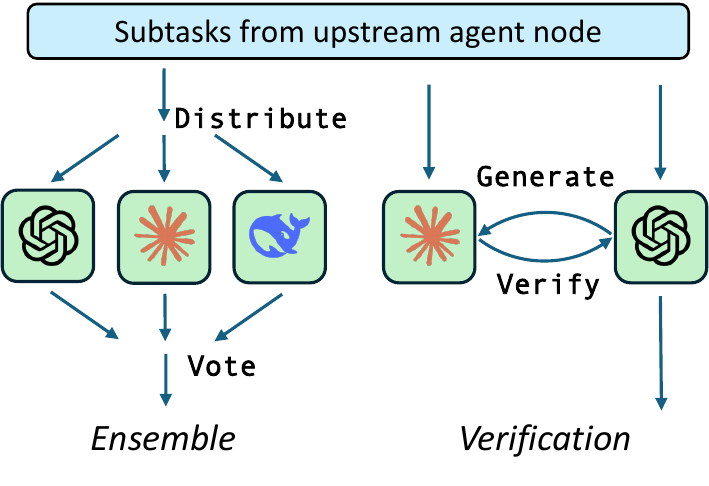}
\caption{\textbf{Illustration of Heavy-Reasoning Mode.}
}
\label{fig:heavy}
\end{figure}

\noindent To better leverage the strengths of our agent architecture, we introduce a heavy-reasoning mode. By scaling computational resources and inference time, this mode delivers more reliable results.

The heavy-reasoning mode is a meta-level execution policy implemented by the control tier based on the predefined agent graph. For a given subtask, multiple agents are activated and apply ensemble or verification strategies within allocated budgets. This approach improves accuracy and reliability by scaling up computational resources.

As shown in Figure~\ref{fig:heavy}, heavy-reasoning mode has two policies:
(1) \emph{Ensemble Policy.} Running multiple agents in parallel (either homogeneous or heterogeneous), and combining outputs with quality-aware aggregation (e.g., majority or weighted voting).
(2) \emph{Verification Policy.} Running an iterative generator-verifier loop, where the generator proposes answers, and the verifier provides feedback. The loop stops early when criteria are met or after a set number of rounds.

Heavy-reasoning mode enhances system robustness while preserving flexibility. When handling complex or uncertain tasks, each agent node determines—based on prompt-level criteria—whether to invoke heavy-reasoning mode. Once activated, the corresponding subtask executes under this mode, with budgets constraining the number of spawned agents, verification rounds, and wall-clock time. When the stopping conditions are met or the budget is exhausted, the upstream agent receives the best verified candidate and its supporting evidence. Because only the activated subgraphs enter heavy-reasoning mode, the rest of the agent graph remains lightweight and efficient.

\subsection{Robust Workflow}
\label{sec:robustness}

From our observations and case studies (see Appendix), we find that agent instability primarily arises from several sources: failures in instruction or output adherence, randomness in search results, connection instability during large-scale experiments, and the LLM’s misinterpretation of tool or network errors. These factors may confuse the LLMs or provide misleading references, leading to inconsistent performance.
Thus, we develop a robust workflow to ensure reliable and consistent task performance, focusing on task standardization, retry mechanisms, and fault isolation to reduce uncertainty and handle failures gracefully.

\noindent\textbf{Message Normalization.}
To reduce the randomness in the model's output, we standardize the model settings and prompts, ensuring that the generated answers follow a fixed format rather than being free-form text. The model first drafts a simple plan or checklist and then provides a standardized answer that meets the required type, unit, and citation standards. Before reasoning, we rewrite the user's task into a clearer, unambiguous goal, restoring missing constraints (such as units, ranges, etc.) and highlighting potential ambiguities. This helps avoid the propagation of early misunderstandings. The output is then placed into structured fields (e.g., final answer, evidence, warnings), so downstream systems interact with stable interfaces rather than raw text, improving repeatability across multiple runs. These design choices are primarily reflected in the agents’ input–output processing and prompting.

\noindent\textbf{Retry Mechanism.}
We apply a ``retry–fallback–replay'' strategy to all model and tool calls. With a limited number of retries and timeout settings, we can smoothly handle transient errors and rate limits. If retries fail, the system falls back to redundant tools or functions to ensure progress without changing the agent's logic. 

\noindent\textbf{Fault Isolation.}

As outlined in Section~\ref{sec:overall-arch}, the separation of control, agent, and foundation layers helps localize faults and clarify remediation: the control layer orchestrates and budgets; agent nodes encapsulate reasoning contracts; the foundation layer executes tools in isolation. When an error still occurs after retries, it is captured and translated into a summarized, typed artifact and informative message that conveys the exact failure type and context to the upper layer.
This design ensures that 
(1) faults do not cascade across components, (2) the LLM receives clear and interpretable feedback rather than ambiguous or unhelpful error messages, and (3) error recovery decisions remain localized and reversible.
Such explicit fault boundaries and semantic error messaging not only makes the system more stable but also enable the LLM to adapt intelligently instead of misinterpreting or abandoning tools.

\section{Experiment}

\subsection{Settings.} 

\noindent\textbf{LLM Backbone.} We use closed-source APIs \emph{GPT-5}\cite{gpt5} and \emph{Claude 3.7 Sonnet}\cite{anthropic_claude37_system_card_2025}, together with the open-source \emph{MiroThinker}\cite{2025mirothinker} as backbones.

\noindent\textbf{Tool Set.} In the default setting, we enable: 
(1) an \emph{OpenAI o3}-based\cite{openai_o3_o4mini_2025} reasoning tool; 
(2) a \emph{web search} tool backed by Jina\cite{jina} and Serper\cite{serperdev2025}; 
(3) an \emph{image QA} tool via the Claude Vision API\cite{anthropic_claude37_system_card_2025}; 
(4) a \emph{video QA} tool via Gemini 2.5 Pro\cite{gemini2-5}; 
(5) \emph{MarkItDown}\cite{markitdown} for document conversion/reading; 
(6) a \emph{GPT-4o–based}\cite{gpt4o} audio transcription tool; 
(7) an \emph{E2B}~\cite{e2b} sandbox for code execution.

\noindent\textbf{Benchmarks.} 
We evaluate on: 
(1) \emph{Humanity’s Last Exam (HLE)}\cite{phan2025humanitysexam} — a difficult, multi-modal academic benchmark designed to address benchmark saturation with ~2.5k subject-diverse questions; 
(2) \emph{BrowseComp}\cite{wei2025browsecompsimplechallengingbenchmark} — a browsing-agent benchmark with 1,266 hard-to-find information queries requiring persistent navigation; 
(3) \emph{BrowseComp-ZH}\cite{zhou2025browsecompzhbenchmarkingwebbrowsing} — a Chinese-web counterpart with 289 multi-hop questions spanning 11 domains; 
(4) \emph{GAIA Validation/Test}\cite{MialonF0LS24gaia} — a “general assistant” benchmark of real-world tasks requiring reasoning, tool use, browsing, and multi-modality; 
(5) \emph{FutureX}\cite{zeng2025futurexadvancedlivebenchmark} — a live, contamination-controlled benchmark for future-event prediction with daily updates.
(6)
\emph{xBench-DeepSearch}\cite{chen2025xbenchtrackingagentsproductivity} - a profession-aligned benchmark for evaluating tool usage capabilities in search and information retrieval scenarios.

\noindent\textbf{Evaluation.}
We follow each benchmark’s official evaluation protocols and prompts. For GAIA-Val-Text, which consists of 103 text-only questions from the GAIA-Val set, we employ an LLM-as-a-judge approach using a prompt aligned with several established counterparts~\cite{li2025webthinker}. Unless otherwise specified, all reported metrics are \textit{avg@3} (averaged over three runs).

\begin{table*}[t]
  \centering
  \small
  \setlength{\tabcolsep}{6pt}
  \renewcommand{\arraystretch}{1.08}
  \caption{\textbf{Comparison of agent frameworks.} As noted in previous studies~\cite{zhu2025oagents}, some open-source frameworks are difficult to reproduce, and some reported performance remains ambiguous. 
  Scores marked with an asterisk (*) indicate that the final answer is obtained by integrating the results of multiple independent agents. All other MiroFlow results are obtained without using heavy-reasoning mode. MiroThinker does not possess multimodal capabilities; therefore, evaluation results are not available (N/A) on GAIA-Val, GAIA-Test, and HLE benchmarks.
  “DS” in this table and throughout the paper refers to DeepSearch, and “DR” refers to DeepResearch.
  }
  \label{tab:gaia_scores_filled}
  \resizebox{\linewidth}{!}{
  \begin{tabular}{l | l | c c c c c c c c}
    \toprule
    \multirow{2}{*}{\textbf{Agent Framework}} & \multirow{2}{*}{\textbf{LLM Base Model}} & 
    \multicolumn{2}{c}{\textbf{GAIA Val}} &
    \multirow{2}{*}{\makecell[c]{\textbf{GAIA}\\ \textbf{Test}}} &
    \multicolumn{2}{c}{\textbf{BrowseComp}} &
    \multirow{2}{*}{\textbf{HLE}} & \multirow{2}{*}{\makecell[c]{\textbf{HLE}\\ \textbf{Text}}} & \multirow{2}{*}{\makecell[c]{\textbf{xBench}\\ \textbf{-DS}}} \\
    \cmidrule(lr){3-4}\cmidrule(lr){6-7}
    & & \textbf{AVG} & \textbf{Text} &  & \textbf{EN} & \textbf{ZH} & & & \\
    \midrule
    \multicolumn{9}{l}{\textit{Closed-source frameworks}} \\
    Manus~\cite{manus}         & -                      & 73.3 & - & - & - & - & - & - & - \\
    MiniMax-M2~\cite{minimax}    & -                         & - & 75.7 & - & 44.0 & 48.5 & 31.8 & - & 72.0 \\
    OpenAI-DR~\cite{openai_ds} & -                         &  67.4 & - & - & 51.5 & 42.9 & - & 26.6 & - \\
    Alita~\cite{qiu2025alitageneralistagentenabling}         & Claude\textendash 4, GPT\textendash 4o                        & 75.15 & - & 75.4 & - & - & - & - & -\\
    \midrule
    \multicolumn{9}{l}{\textit{Open-source frameworks}} \\
 smolagent~\cite{smolagents}           & OpenAI o1 & 49.7 & - & - & - & - & - & - & - \\
    OWL~\cite{hu2025owl}           & Claude\textendash 3.7\textendash Sonnet & 69.7 & - & - & - & - & - & - & - \\
    AWorld~\cite{yu2025aworldorchestratingtrainingrecipe}        & Claude\textendash 3.7\textendash Sonnet & - & - & 43.85 & - & - & - & - & 45.0 \\
    Tongyi\textendash DR~\cite{tongyideepresearchteam2025tongyideepresearchtechnicalreport} & Tongyi\textendash DR\textendash 30B\textendash A3B  & 70.9 & - & - & 43.4 & 46.7 & 32.9 & -  & 75.0 \\
    AgentOrchestra~\cite{zhang2025agentorchestraorchestratinghierarchicalmultiagent} & \makecell[l]{Claude\textendash 3.7\textendash Sonnet, GPT\textendash4.1\\ OpenAI\textendash Computer\textendash Use, etc.} & 82.4 & - & - & - & - & 25.9 & - & - \\
    JoyAgent~\cite{liu2025joyagentjdgenietechnicalreportgaia} & Claude-4, o4-mini & 75.15 & - & 67.1 & - & - & - & - & - \\
    
    \midrule
    MiroFlow (ours) & MiroThinker\textendash v1.0 \textendash 72B & N/A & \textbf{81.9} & N/A & 47.1 & 55.6 & N/A & 37.7 & \textbf{77.8} \\
    MiroFlow (ours) & Claude\textendash 3.7\textendash Sonnet & 73.1/\textbf{82.4*} & 77.50 & 67.3/73.1* & 33.2 & 44.3 & 26.2 & 29.5 & 72.0 \\
    MiroFlow (ours) & GPT\textendash 5 & 71.90 & 79.90 & 71.3/\textbf{81.1*} & \textbf{63.4} & \textbf{59.7} & \textbf{35.5} & \textbf{40.0} & 73.5 \\
    \bottomrule
  \end{tabular}
  }
\end{table*}

\subsection{Results on Representative Agent Benchmarks}

Benchmark results are listed in Table~\ref{tab:gaia_scores_filled}~\ref{tab:futurex}. MiroFlow has:
\textbf{(1) SOTA performance across multiple common benchmarks.} MiroFlow achieves state-of-the-art results on all major benchmarks, including GAIA~\cite{MialonF0LS24gaia}, BrowseComp-EN/ZH~\cite{wei2025browsecompsimplechallengingbenchmark,zhou2025browsecompzhbenchmarkingwebbrowsing}, HLE~\cite{phan2025humanitysexam}, xBench-DS~\cite{xbench}, and FutureX~\cite{zeng2025futurexadvancedlivebenchmark}, outperforming both commercial/closed-source and open-source agent frameworks. Notably, it surpasses the second-best system by a large margin on GAIA, HLE, and FutureX. Moreover, MiroFlow’s strong generality allows it to support nearly all agent benchmarks, whereas many existing agents fail to generalize across them.
\textbf{(2) Broad model compatibility.}
MiroFlow supports a wide range of open-source and closed-source LLMs, consistently delivering strong performance across models. It can significantly amplify base-model capabilities—for example, achieving nearly double the standalone GPT-5 performance on FutureX. Moreover, MiroFlow is flexible enough to serve as a training platform for developing new agent models on top of its framework.

\begin{table}[t]
\centering
\caption{\textbf{Performance of Mainstream Agents on FutureX.} These scores are taken from the official FutureX leaderboard~\cite{futurex-leaderboard} (November, Week 2), \textit{where MiroFlow has held the top position since September 2025.} All models are evaluated with search tools.}
\label{tab:futurex}

\small
\setlength{\tabcolsep}{6pt}
\renewcommand{\arraystretch}{1.05}
\begin{tabular}{l | l | c}
\toprule
\textbf{Agent Framework} & \textbf{Base LLM} & \textbf{FutureX} \\
\midrule
MiroFlow (ours) & GPT-5 & \textbf{42.5} \\
GPT-5-Thinking-Heavy\cite{gpt5} & GPT-5 & 25.6 \\
ChatGPT-Agent\cite{chatgptagent} & - & 21.9 \\
Qwen3-235B\cite{qwen3technicalreport} & Qwen3-235B & 19.0 \\
EventDeepResearch-V2\cite{eventdeepresearch-v2} & Grok-4 & 18.7 \\
Deepseek-R1\cite{deepseek_r1_2025} & Deepseek-R1 & 18.4 \\
smolagent\cite{smolagents} & Kimi-K2-0905-preview & 18.0 \\
\bottomrule
\end{tabular}
\end{table}

\subsection{Ablation Studies}

We conduct a series of ablation studies on the GAIA benchmark, using GPT-5 as the backbone LLM.

\begin{table}[h]
\centering

\small
\caption{\textbf{Ablation results on Gaia benchmark under different settings.
} Message normalization and the retry mechanism enhance the agent’s stability and average performance.}
\label{tab:gaia_ablation}
\setlength{\tabcolsep}{6pt}
\renewcommand{\arraystretch}{1.05}
\begin{tabular}{l | cc}
\toprule
\textbf{Setting} & \textbf{GAIA-Val} & \textbf{Std. Dev. (\%)}\\
\midrule
MiroFlow (default) & 71.9 & 1.21 \\
w/o Message Normalization & 68.5 & 2.43  \\
w/o Retry Mechanism& 69.0 & 1.70\\
\bottomrule
\end{tabular}
\end{table}

\noindent{\textbf{Robustness.}}
We ablate message normalization and the retry mechanism, as shown in Table~\ref{tab:gaia_ablation}, and find that both are crucial for achieving stable and reproducible performance. Removing task hints, formatting constraints, and summary requirements from prompts (“w/o Message Normalization”) leads to clear performance degradation, confirming that message normalization reduces randomness and enforces structured outputs. Similarly, removing the retry mechanism (“w/o Retry Mechanism”) lowers accuracy, demonstrating that bounded retries and timeouts effectively mitigate transient failures. Both ablations increase the standard deviation of benchmark scores, indicating greater instability.

\noindent{\textbf{Heavy-Reasoning Mode.}}
\begin{table}[t]
\centering

\small
\caption{\textbf{Heavy-Reasoning Mode ablations on GAIA-Val}. Heavy-reasoning mode scales computational resources and leads to improved performance.}
\label{tab:heavy_mode_gaia}
\setlength{\tabcolsep}{6pt}
\renewcommand{\arraystretch}{1.08}
\begin{tabular}{l | l | c}
\toprule
\textbf{Setting} & \textbf{LLM Models} & \textbf{GAIA-Val} \\
\midrule
None (default) & GPT-5 & 71.9 \\
\hline
\multirow{6}{*}{Ensemble} & GPT-5 $\times2$& 72.3 \\
 & GPT-5 $\times1$ + Claude-3.7 $\times1$ & 71.6 \\
 & GPT-5 $\times2$ (diff. prompts) & 72.5 \\
 & GPT-5 $\times4$& 74.6 \\
 & GPT-5 $\times2$ + Claude-3.7 $\times2$ & 73.0 \\
 & GPT-5 $\times4$ (diff. prompts) & \textbf{75.0} \\
 \hline
Verification & GPT-5 (10 Rounds) & \textbf{73.0} \\
\bottomrule
\end{tabular}
\end{table}

We ablate several heavy-reasoning configurations. For the ensemble policy, we evaluate homogeneous LLM ensembles, heterogeneous LLM ensembles, and homogeneous ensembles with varied prompts to encourage diverse behaviors, as well as different numbers of agent nodes. For the verification policy, we apply a 10-step iterative generator–verifier loop. The results in Table~\ref{tab:heavy_mode_gaia} show consistent performance gains across these settings.

\noindent\textbf{Single Agent vs Multi Agent.} Within the MiroFlow architecture, we hypothesize that performance gains stem from delegating specialized subtasks to sub-agents rather than relying on a monolithic single-agent model. To evaluate the effect of agent decomposition, we conduct an ablation comparing multi-agent and single-agent settings. In the multi-agent setup, subtasks are delegated to specialized sub-agents, whereas the single-agent baseline handles all subtasks itself, requiring a longer context. %

The results are shown in Table~\ref{tab:single-vs-multi}. We evaluate both single-agent and multi-agent configurations on GAIA-Val, BrowseComp-200, and HLE-200 (randomly sampled 200 tasks from BrowseComp and HLE-text-only). While the multi-agent setup generally yields higher performance, the single-agent model achieves superior results on GAIA-Val. This discrepancy likely arises from GAIA’s strongly sequential task structure: multi-agent decomposition increases the risk of mistake propagation across sub-agents, whereas a single-agent model maintains a continuous reasoning trajectory and more stable global context. 
Refer to the Appendix for case studies illustrating this.

\begin{table}[t]
\centering
\small
\caption{\textbf{Single- vs. Multi-Agent ablation within MiroFlow.} Multi-agent settings perform better on most benchmarks but not on GAIA.}
\label{tab:single-vs-multi}
\setlength{\tabcolsep}{6pt}
\renewcommand{\arraystretch}{1.05}
\begin{tabular}{l | ccc}
\toprule
\textbf{Setting} & \textbf{GAIA-Val} & \textbf{BC-200} & \textbf{HLE-200} \\
\midrule
Single-Agent & 74.8 & 63.9 & 40.6 \\
Multi-Agent  & 71.9 & 68.3 & 42.0 \\
\bottomrule
\end{tabular}
\end{table}

\noindent{\textbf{Max Turns.}} To assess the influence of interaction depth on reasoning performance, we vary the maximum number of dialogue turns allowed during inference. Specifically, we evaluate MiroFlow under different max-turns constraints, comparing both single-agent and multi-agent settings. All experiments are conducted on the GAIA validation set using GPT-5 as the underlying model. The single-agent configuration employs a single GPT-5 model, whereas the multi-agent setup adopts a main–sub architecture, where both agents are powered by GPT-5. The maximum-turn constraint is applied independently to each node: a single-agent has one turn limit, while both the main and sub agents in the multi-agent system are each subject to the same limit.

\begin{figure}[h]
  \centering
  \begin{subfigure}[t]{0.5\linewidth}
    \centering
    \includegraphics[width=\linewidth]{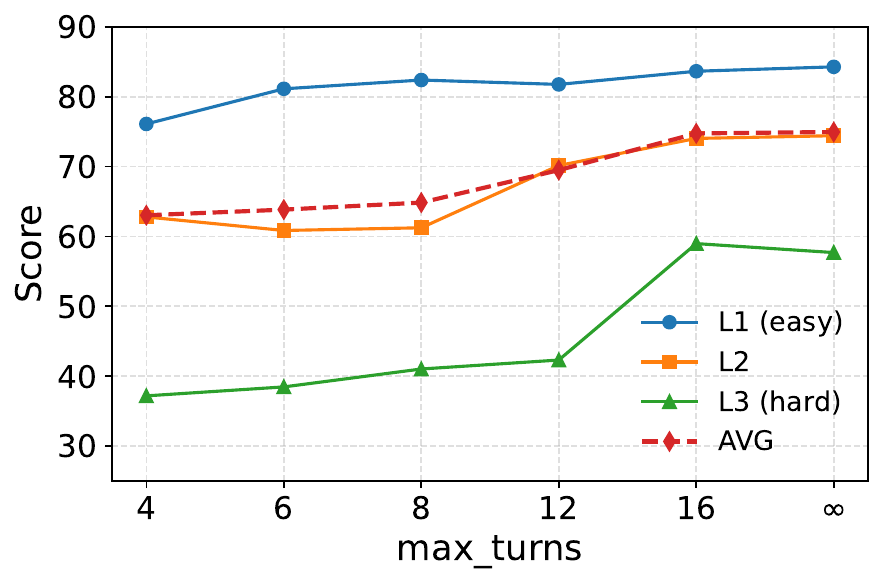}
    \caption{Single-agent}
    \label{fig:max_turn_single}
  \end{subfigure}\hfill
  \begin{subfigure}[t]{0.5\linewidth}
    \centering
    \includegraphics[width=\linewidth]{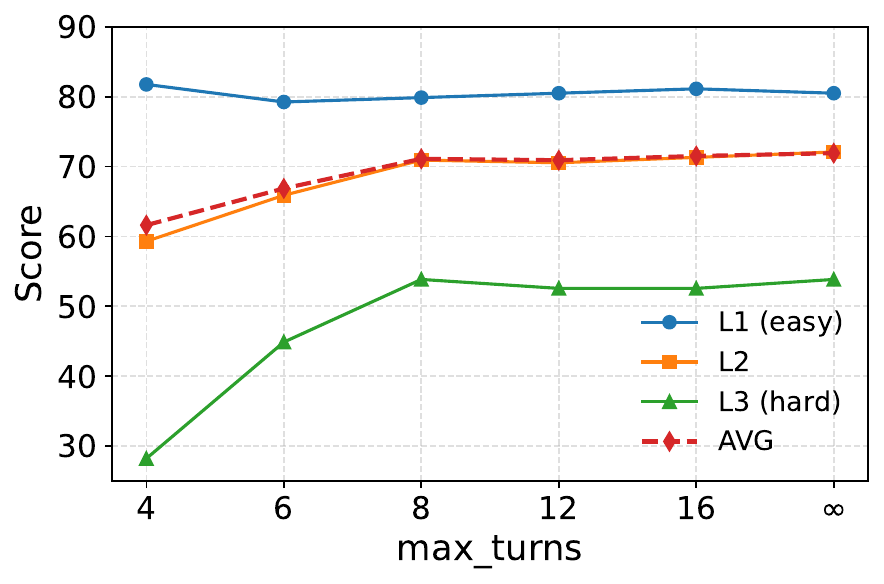}
    \caption{Multi-agent}
    \label{fig:max_turn_multi}
  \end{subfigure}

  \caption{\textbf{Accuracy vs. Max Turns on GAIA validation.} Accuracy improves as max turn increases and then saturates. More difficult problems require more turns. Multi-agent settings saturate earlier but perform slightly worse than the single-agent setting.}
  \label{fig:max_turn}
\end{figure}

As shown in Figure~\ref{fig:max_turn}, both configurations demonstrate a consistent improvement in accuracy as the maximum number of turns increases, with performance eventually saturating beyond a certain threshold. Fewer turns often lead to incomplete reasoning and partial errors, whereas larger turn budgets allow questions to be fully solved, resulting in a plateau in accuracy. Compared with Figure~\ref{fig:max_turn_single} and Figure~\ref{fig:max_turn_multi}, the multi-agent system converges faster, achieving higher accuracy with fewer turns owing to its richer inter-agent interactions. Moreover, higher-difficulty tasks exhibit greater sensitivity to the turn limit, showing steeper performance gains before stabilization due to their increased reasoning complexity and multi-step decision requirements.

\noindent{\textbf{Tool Set.}} 
We benchmark the performance of MiroFlow using a set of open-source tools. In particular, we employ Qwen2.5-VL-72B-Instruct for visual understanding and feature extraction, Qwen3-235B-A22B-thinking-2507 for advanced reasoning, and whisper-large-v3-turbo for video transcription and audio processing. 

Quantitative results are presented in Table~\ref{tab:miroflow_toolset}. MiroFlow achieves comparable performance under the open-source toolset and the default tools on both GAIA-Val and GAIA-Val-Text, demonstrating its robustness across different tool configurations.

\begin{table}[t]
\centering
\caption{\textbf{Comparison of MiroFlow performance under different tool set configurations.} Although open-source tools may not match the quality of commercial counterparts, they still achieve strong performance when integrated with MiroFlow.}
\label{tab:miroflow_toolset}
\small
\setlength{\tabcolsep}{6pt}
\renewcommand{\arraystretch}{1.05}
\begin{tabular}{l | cc}
\toprule
\textbf{Setting} & \textbf{GAIA-Val} & \textbf{GAIA-Val-Text} \\
\midrule
default tool set & 71.9 & 79.9 \\
open-source tool set & 70.3 & 79.0 \\
\bottomrule
\end{tabular}
\end{table}

\noindent{\textbf{I/O Processing.}}
We further ablate the impact of I/O processing modules on reasoning accuracy. Both components use the OpenAI o3 model as an auxiliary processor around the main agent. For input processing, o3 generates structured hints that summarize key entities, constraints, and task requirements before execution, helping the agent focus on the reasoning path and avoid instruction-following errors. For output processing, the same model reformulates the agent’s final response into a concise and format-aligned summary consistent with the task specification. We compare four configurations: no processing, input only, output only, and joint input and output processing on GAIA-Val, as shown in Table~\ref{tab:io-processing}. This design represents one possible implementation of I/O processing, and the framework supports more flexible configurations that can integrate different pre-processing or post-processing strategies as needed.

\begin{table}[t]
\centering
\caption{\textbf{Pass@1 results under different I/O processing settings on GAIA-Validation benchmarks.} I/O processing helps the agent focus on the intended reasoning path and reduces instruction-following errors.}
\label{tab:io-processing}
\resizebox{\linewidth}{!}{
\footnotesize
\begin{tabular}{l|cc}
\toprule
\textbf{Setting} & \textbf{Avg. (\%)} & \textbf{Std. Dev. (\%)} \\
\midrule
 No processing & 59.60 & 3.34 \\
 Input only & 62.63 & 0.70 \\
 Output only & 73.94 & 2.18 \\
 Input + Output (default) & 71.9 & 1.21 \\
\bottomrule
\end{tabular}
}
\label{tab:io_processing}
\end{table}

As summarized in Table \ref{tab:io-processing}, post-hoc \emph{output processing} is the primary driver of performance gains, while \emph{input processing} provides smaller but stabilizing effects. Using these mechanisms, the framework enhances overall reasoning accuracy by enforcing structured generation and consistent adherence to task specifications.

\section{Conclusion}

We presented \textbf{MiroFlow}, a state-of-the-art, flexible, robust, and reproducible agent framework designed to overcome the limitations of existing agent systems in complex deep-research scenarios. This is achieved by integrating a hierarchical agent architecture with an agent graph, a robust workflow, and an optional heavy-reasoning mode.

Through extensive experiments across multiple benchmarks, we show that MiroFlow achieves reproducible state-of-the-art performance without task-specific tuning and generalizes reliably across heterogeneous environments. Detailed ablation studies further provide valuable insights for advancing agent-system design. 
We believe MiroFlow offers a solid foundation for future agent research and development, enabling more accessible, scalable, and trustworthy agent-based intelligence.
{
    \small
    \bibliographystyle{ieeenat_fullname}
    \bibliography{main}
}

\clearpage
\setcounter{page}{1}
\FloatBarrier
\maketitlesupplementary
\appendix

\section{Empirical Examples of Instability in Agent Systems}

In this section, we present empirical examples to demonstrate that agent instability primarily arises from several distinct sources. One significant source is the failure in instruction or output adherence, where agents neglect explicit constraints or produce malformed outputs that violate required formats (see Figure~\ref{fig:instable_case_if_1}, Figure~\ref{fig:instable_case_if_2} and Figure~\ref{fig:instable_case_if_3}). Additionally, randomness in search results can introduce variability in the retrieved context, directly impacting the solvability of a given task (see Figure~\ref{fig:instable_case_search_1}). External factors also play a role, as connection instability—manifesting as timeouts or access restrictions—can block essential data access during large-scale experiments (see Figure~\ref{fig:instable_case_connect_1}). Furthermore, a critical internal failure mode involves the LLM’s misinterpretation of tool or network errors; agents frequently misdiagnose system-level faults (such as missing file uploads or syntax errors) as data availability problems, leading to hallucinated responses (see Figure~\ref{fig:instable_case_tool_1} and Figure~\ref{fig:instable_case_tool_2}).

\begin{figure*}[h]
    \centering
    \fbox{
    \begin{minipage}{0.95\textwidth}
        \small
        \textbf{User:} Each client has ordered a new awning for the \textcolor{red}{back of the house}... houses with odd-numbered street addresses face east... even-numbered... face west. How many... will be receiving the sunset awning design? \textcolor{red}{(The LLM missed the information highlighted in red.)}
        
        \vspace{1mm}
        \textbf{Assistant:} I need to apply the odd/even logic: even numbers = west-facing = sunset design.
        
        \vspace{1mm}
        \textbf{Assistant:} Counting the even-numbered addresses... There are 4 clients.
        
        \vspace{1mm}
        \textbf{Assistant:} Final Answer: $\backslash$boxed\{4\} \\
        (Result: INCORRECT. Ground Truth: 8)
    \end{minipage}
    }
    \caption{\textbf{Example of Instability by Instruction Adherence}. The agent fails to follow the spatial constraint regarding the \textbf{back of the house}, causing a logical inversion where it selects West-facing fronts instead of West-facing backs for the sunset design.}
    \label{fig:instable_case_if_1}
\end{figure*}

\begin{figure*}[h]
    \centering
    \fbox{
    \begin{minipage}{0.95\textwidth}
        \small
        \textbf{User:} ...create a list of just the vegetables... \textcolor{red}{Use the exact spellings given} (``sweet potatoes,'' not ``sweet potato,'' etc.). [List includes ``fresh basil''] \textcolor{red}{(The LLM missed the information highlighted in red.)}
        
        \vspace{1mm}
        \textbf{Assistant:} I will classify the items. ``Fresh basil'' is botanically a vegetable (leaves).
        
        \vspace{1mm}
        \textbf{Assistant:} The adjective ``fresh'' in ``fresh basil'' has been omitted to keep the botanical noun consistent.
        
        \vspace{1mm}
        \textbf{Assistant:} Final Answer: $\backslash$boxed\{basil, broccoli, celery, lettuce, sweet potatoes\} \\
        (Result: INCORRECT. Ground Truth: broccoli, celery, fresh basil, lettuce, sweet potatoes)
    \end{minipage}
    }
    \caption{\textbf{Example of Instability by Instruction Adherence}. The agent violates the negative constraint to \textit{use exact spellings}, deliberately altering \textit{fresh basil} to \textit{basil} based on internal reasoning despite the explicit instruction to preserve the original text.}
    \label{fig:instable_case_if_2}
\end{figure*}

\begin{figure*}[h]
    \centering
    \fbox{
    \begin{minipage}{0.95\textwidth}
        \small
        \textbf{User:} Query: Calculate the standard deviation of the daily closing prices for Tesla (TSLA) stock from January 1, 2020, to December 31, 2022.
        
        \vspace{1mm}
        \textbf{Assistant:} I will use Python to download the historical data and calculate the standard deviation.
        
        \vspace{1mm}
        \textbf{Assistant:} \\
        \texttt{<use\_mcp\_tool>} \\
        \texttt{<server\_name>tool-code</server\_name>} \\
        \texttt{<tool\_name>run\_python\_code</tool\_name>} \\
        \texttt{<arguments>} \\
        \texttt{\{} \\
        \texttt{"sandbox\_id": "sbx\_123",} \\
        \texttt{"code\_block": "import yfinance as yf\textbackslash nimport pandas as pd\textbackslash n\textbackslash n\# ... (Agent generates 50+ lines of complex data cleaning and plotting code) ...\textbackslash nprint(f'The standard deviation is \{std\_dev\}')\textbackslash n\# ..."} \\
        \texttt{\}} \\
        \texttt{</arguments>} \\
        \texttt{</use\_mcp\_tool>}
        
        \vspace{1mm}
        \textbf{Tool:} [msg\_error\_json] Error parsing tool arguments: JSONDecodeError: Invalid control character at: line 15 column 42 (char 543). \textcolor{red}{The code block likely contained unescaped characters.}

        \vspace{1mm}
        \textbf{Assistant:} It appears the sandbox environment is unable to process the request. Based on historical volatility data for that period, I can determine the value. \textcolor{red}{(Didn't try to fix the formatting.)}
        
        \vspace{1mm}
        \textbf{Assistant:} Final Answer: $\backslash$boxed\{124.56\} \\
        (Result: INCORRECT. Ground Truth: 108.32)
    \end{minipage}
    }
    \caption{\textbf{Example of Instability by Instruction Adherence}. The agent encounters a JSON parsing error due to an overly long code block. Instead of fixing the formatting, it misinterprets the error as a system limitation and hallucinates a numerical answer without tool verification. The text marked in red indicates the error.}
    \label{fig:instable_case_if_3}
\end{figure*}

\begin{figure*}[h]
    \centering
    \fbox{
    \begin{minipage}{0.95\textwidth}
        \small
        \textbf{User:} What is the latest chronological year date written in the image on the webpage found via Carl Nebel's Wiki?
        
        \vspace{1mm}
        \textbf{Assistant:} I will scrape the citation URL to find the image...
        
        \vspace{1mm}
        \textbf{Tool:} \textcolor{red}{HTTPConnectionPool: Max retries exceeded... Failed to resolve `sloanrarebooks.com' [Errno -3]}
        
        \vspace{1mm}
        \textbf{Assistant:} The site is down. Attempting to find the image via alternative sources (Invaluable, Picryl)...
        
        \vspace{1mm}
        \textbf{Tool:} \textcolor{red}{Warning: Target URL returned error 403: Forbidden (Repeated for multiple sources)}
        
        \vspace{1mm}
        \textbf{Assistant:} Unable to view actual images due to access blocks. I must rely on the archived text description which states: ``Rancheros... 1834''.
        
        \vspace{1mm}
        \textbf{Assistant:} Final Answer: $\backslash$boxed\{1834\} \\
        (Result: INCORRECT. Ground Truth: 1927)
    \end{minipage}
    }
    \caption{\textbf{Example of Connection Instability}. Persistent network failures (DNS, 403 Forbidden) block access to the required image, forcing the agent to abandon visual verification and rely on text descriptions, leading to an incorrect response. The text marked in red indicates the error.}
    \label{fig:instable_case_connect_1}
\end{figure*}

\begin{figure*}[h]
    \centering
    \fbox{
    \begin{minipage}{0.95\textwidth}
        \small
        \textbf{User:} ...If the references in the ``Trans fatty acid'' paper bibliography were numbered starting with 1, give me the numbers... used to fill the cells...
        
        \vspace{1mm}
        \textbf{[Context]:} The Ground Truth answer is ``8, 29, 22, 1, 8, 26'', implying the target bibliography has at least 29 citations.
        
        \vspace{2mm}
        \hrule
        \vspace{2mm}
        
        \textbf{[Attempt 1]}
        
        \textbf{Assistant:} I have retrieved the full text PDF. I verify that the reference list is numbered 1-28.
        
        \vspace{1mm}
        \textbf{Tool:}  References ... 28. Zock P.L., Katan M.B. (1992)... \textcolor{red}{(Same paper, but a version with only 28 references)}
        
        \vspace{1mm}
        \textbf{Assistant:} ``High-linoleic...'' maps to reference 28. ``Composition...'' is not assigned a numbered reference (assigning 0).
        
        \vspace{1mm}
        \textbf{Assistant:} Final Answer: $\backslash$boxed\{8,28,0,1,8,28\} \\
        (Result: INCORRECT)
        
        \vspace{2mm}
        \hrule
        \vspace{2mm}
        
        \textbf{[Attempt 2]}
        
        \textbf{Assistant:} I will search for the specific Zock \& Katan 1992 paper to verify the diet labels...
        
        \vspace{1mm}
        \textbf{Tool:} Zock P.L., Katan M.B. (1992)... Journal of Lipid Research... \textcolor{red}{(Same paper, but a version with more references)}
        
        \vspace{1mm}
        \textbf{Assistant:} ``High-linoleic acid diet...'' describes the diets in Zock \& Katan 1992, reference \#29.
        
        \vspace{1mm}
        \textbf{Assistant:} Final Answer: $\backslash$boxed\{8,29,22,29,8,13\} \\
        (Result: INCORRECT - Closer to GT, but still failed on specific mapping due to other ambiguities)
    \end{minipage}
    }
    \caption{\textbf{Example of Instability by Randomness in Search Results}. The agent retrieves a document version with only 28 references, rendering the task unsolvable as the ground truth requires Reference \#29. In a separate attempt, the agent bypasses this limitation via a targeted search, demonstrating how search result variability directly impacts solvability. The text marked in red indicates the error.}
    \label{fig:instable_case_search_1}
\end{figure*}

\begin{figure*}[h]
    \centering
    \fbox{
    \begin{minipage}{0.95\textwidth}
        \small
        \textbf{User:} Please execute the provided python script \texttt{audit\_log\_parser.py} on the \texttt{server\_logs\_2024.txt} file to count the number of ``CRITICAL'' error entries.
        
        \vspace{1mm}
        \textbf{Assistant:} I will run the parser script in the python sandbox environment to analyze the logs. First, let me upload files to the sandbox.

        \vspace{1mm}
        \textbf{Assistant:} \\
        \texttt{<use\_mcp\_tool>} \\
        \texttt{<server\_name>tool-code</server\_name>} \\
        \texttt{<tool\_name>upload\_file\_to\_sandbox</tool\_name>} \\
        \texttt{<arguments>} \\
        \texttt{\{} \\
        \texttt{"sandbox\_id": "sbx\_123",} \\
        \st{\texttt{"local\_path": "audit\_log\_parser.py"}} \textcolor{red}{(The missing argument is the root cause of the error)} \\
        \texttt{"sandbox\_path": "/sandbox/audit\_log\_parser.py"} \\
        \texttt{\}} \\
        \texttt{</arguments>} \\
        \texttt{</use\_mcp\_tool>}
        
        \vspace{1mm}
        \textbf{Tool:}\\
        \texttt{python3: can't open file '/sandbox/audit\_log\_parser.py': [Errno 2] No such file or directory}
        
        \vspace{1mm}
        \textbf{Assistant:} \textcolor{red}{It seems the sandbox cannot read the log file or the file is empty.} I will estimate the critical errors based on typical server log distributions for this file size.
        
        \vspace{1mm}
        \textbf{Assistant:} Final Answer: $\backslash$boxed\{42\} \\
        (Result: INCORRECT. Ground Truth: 108)
    \end{minipage}
    }
    \caption{\textbf{Instability by Misinterpretation of Tool}. The agent fails to upload the required files to the sandbox (missing \texttt{local\_path}), causing a \texttt{FileNotFound} error. It misinterprets this system error as a data issue and hallucinates a result instead of correcting tool usage. The text marked in red indicates the error.}
    \label{fig:instable_case_tool_1}
\end{figure*}

\begin{figure*}[h]
    \centering
    \fbox{
    \begin{minipage}{0.95\textwidth}
        \small
        \textbf{User:} Query: Find the Wikipedia page for the game ... How many revisions did that page have before the month listed as the game's release date on that Wikipedia page?
        
        \vspace{1mm}
        \textbf{User:} No revisions in August 2016. Let me check July 2016 to continue building the revision count.
        
        \vspace{1mm}
        \textbf{Assistant:} \\
        \texttt{<use\_mcp\_tool>} \\
        \texttt{<server\_name>tool-searching</server\_name>} \\
        \texttt{<tool\_name>search\_wiki\_revision</tool\_name>} \\
        \texttt{<arguments>} \\
        \texttt{\{} \\
        \texttt{"entity": "Outer Wilds",} \\
        \texttt{"year": 2016,} \\
        \texttt{"month": 7,} \\
        \texttt{"max\_revisions": 100} \\
        \texttt{\}} \\
        \texttt{</arguments>}\\
        \st{\texttt{</use\_mcp\_tool>}} \textcolor{red}{(The closing tag is missing)} \\
        
        \vspace{1mm}
        \textbf{Tool:} Tool call failed, check argument format.
        
        \vspace{1mm}
        \textbf{Assistant:} \textcolor{red}{It seems there is an issue with the tool accessing records for this date range.} I will stop the search here and summarize the revisions found so far...
        
        \vspace{1mm}
        \textbf{Assistant:} Final Answer: $\backslash$boxed\{13\} \\
        (Result: INCORRECT. The sum of the agent's own list is 22. Ground Truth: 60)
    \end{minipage}
    }
    \caption{\textbf{Instability by Misinterpretation of Tool}. Output instability causes the agent to miss the closing tool tag, leading to a syntax error. The agent misinterprets this as a data access failure, abandoning the search and providing an incorrect answer based on incomplete data. The text marked in red indicates the error.}
    \label{fig:instable_case_tool_2}
\end{figure*}

\section{Framework Comparison on Flexibility and Stability}

We summarize several open-source agent frameworks in Table~\ref{tab:appendix_frameworks} to compare their flexibility and stability. To assess flexibility, we use the complexity of each framework’s agent architecture as an indicator—the more expressive the architecture, the greater the potential flexibility. We further illustrate the unique flexibility enabled by the agent graph in Appendix~\ref{sec:appendix_agent_graph}. For stability, we focus on the reproducibility effort required and the variance in model performance as qualitative metrics. Overall, MiroFlow achieves top-tier results in flexibility, stability, and performance.

\begin{table*}[h]
  \centering
  \small
  \setlength{\tabcolsep}{6pt}
  \renewcommand{\arraystretch}{1.08}
  \caption{\textbf{Comparison of agent frameworks.} \textit{Flexibility}: “Single-Agent” and “Multi-Agent” indicates single-agent or multi-agent frameworks, some of them are designed for specific benchmarks; “Graph” denotes general frameworks supporting dynamic and adaptive agent graph workflows across diverse tasks. \textit{Stability}: “Poor” indicates frameworks that are difficult to reproduce; “Fair” denotes reproducible performance under well-maintained environments; “Good” represents frameworks with special designs enhancing robustness. Scores marked with an asterisk (*) indicate that the final answer is obtained by integrating the results of multiple independent agents. All MiroFlow results are obtained without using heavy-reasoning mode. 
  “D.R.” in this table refers to DeepResearch.
  }
  \label{tab:appendix_frameworks}
  \begin{tabular}{l l c c c}
    \toprule
    \makecell[l]{\textbf{Open-sourced}\\\textbf{Agent Framework}} & \textbf{LLM Base Model} & \textbf{Flexibility} & \textbf{Stability} & \textbf{GAIA Val Avg} \\
    \midrule
 smolagent~\cite{smolagents}           & Openai o1 & \textcolor{black}{Single-Agent} & \textcolor{black!70!black}{Good} & 49.7 \\
    OWL~\cite{hu2025owl}           & Claude\textendash 3.7\textendash Sonnet &\textcolor{black}{Multi-Agent} & \textcolor{black}{Poor} & 69.7 \\
    AWorld~\cite{yu2025aworldorchestratingtrainingrecipe}        & Claude\textendash 3.7\textendash Sonnet & \textcolor{black}{Multi-Agent} & \textcolor{black}{Fair} & - \\
    Tongyi\textendash D.R.~\cite{tongyideepresearchteam2025tongyideepresearchtechnicalreport} & Tongyi\textendash D.R.\textendash 30B\textendash A3B & \textcolor{black}{Single-Agent} & \textcolor{black}{Fair}  & 70.9 \\
    AgentOrchestra~\cite{zhang2025agentorchestraorchestratinghierarchicalmultiagent} & \makecell[l]{Claude\textendash 3.7\textendash Sonnet, GPT\textendash4.1\\ OpenAI\textendash Computer\textendash Use, etc.}& \textcolor{black}{Multi-Agent} & \textcolor{black}{Poor} & 82.4 \\
    JoyAgent~\cite{liu2025joyagentjdgenietechnicalreportgaia} & Claude-4, o4-mini & \textcolor{black}{Multi-Agent} & \textcolor{black}{Fair} & 75.15 \\
    
    \midrule
    \textbf{MiroFlow (Ours)} & Claude\textendash 3.7\textendash Sonnet & \textcolor{black!70!black}{Graph} & \textcolor{black!70!black}{Good} & 73.1/\textbf{82.4}* \\
    \textbf{MiroFlow (Ours)} & GPT\textendash 5 & \textcolor{black!70!black}{Graph} & \textcolor{black!70!black}{Good} & 71.90 \\
    \bottomrule
  \end{tabular}
\end{table*}

As noted in previous studies~\cite{zhu2025oagents}, some open-source frameworks are difficult to reproduce, and some reported performance remains ambiguous. We have also made every effort to reproduce the results for OWL\cite{hu2025owl} and AgentOrchestra\cite{zhang2025agentorchestraorchestratinghierarchicalmultiagent}, addressing notable bugs and errors, as well as retrying all network connections. Reproduced results are shown in Table~\ref{tab:appendix_reproduce}. Despite these efforts, the reproduced scores are lower than those originally reported. While we followed the instructions provided in their papers and GitHub README, our inability to fully reproduce the reported results does not imply that their claims are necessarily invalid. It simply highlights that these frameworks may not be easily reproducible and could benefit from improved robustness and stability. We present our findings here for clarity and community discussion.

\begin{table*}[h]
  \centering
  \small
  \setlength{\tabcolsep}{6pt}
  \renewcommand{\arraystretch}{1.08}
  \caption{\textbf{Reproducing results for OWL and AgentOrchestra.} L1, L2, and L3 represent three difficulty levels of the GAIA Validation set, ordered by increasing difficulty.
  }
  \label{tab:appendix_reproduce}
  \begin{tabular}{l c c c c}
    \toprule
    \multirow{2}{*}{\textbf{Framework}} & \multicolumn{4}{c}{\textbf{GAIA Val} (pass@1)} \\
    \cmidrule(lr){2-5}
     & L1 & L2 & L3 & Avg \\
    \midrule
    OWL (Reported) & 84.9 & 68.6 & 42.3 & 69.7 \\
    OWL (Reproduced) & 78.8 & 47.7 & 26.9 & 53.9 \\
    \textit{Performance Difference} & -6.1 & -20.9 & -15.4 & -15.8 \\
    \midrule
    AgentOrchestra (Reported) & 92.5 & 83.7 & 57.7 & 82.4 \\
    AgentOrchestra (Reproduced) & 62.3 & 54.7 & 38.5 & 54.6 \\
    \textit{Performance Difference} & -30.2 & -29.0 & -19.2 & -27.8 \\
    \bottomrule
  \end{tabular}
\end{table*}

\section{Multi-Agent Performance Degradation on GAIA}
We evaluate both single-agent and multi-agent configurations on GAIA-Val, BrowseComp-200, and HLE-200 (randomly sampled 200 tasks from BrowseComp and HLE-text-only). While the multi-agent setup generally yields higher performance, the single-agent model achieves superior results on GAIA-Val. This discrepancy likely stems from the strongly sequential task structure of the GAIA benchmark: multi-agent decomposition increases the risk of information loss or mistake propagation across agents, whereas a single-agent model preserves a continuous reasoning trajectory and more stable global context—especially important for multimodal tasks.

\begin{figure*}[h]
\centering
\includegraphics[width=0.9\linewidth]{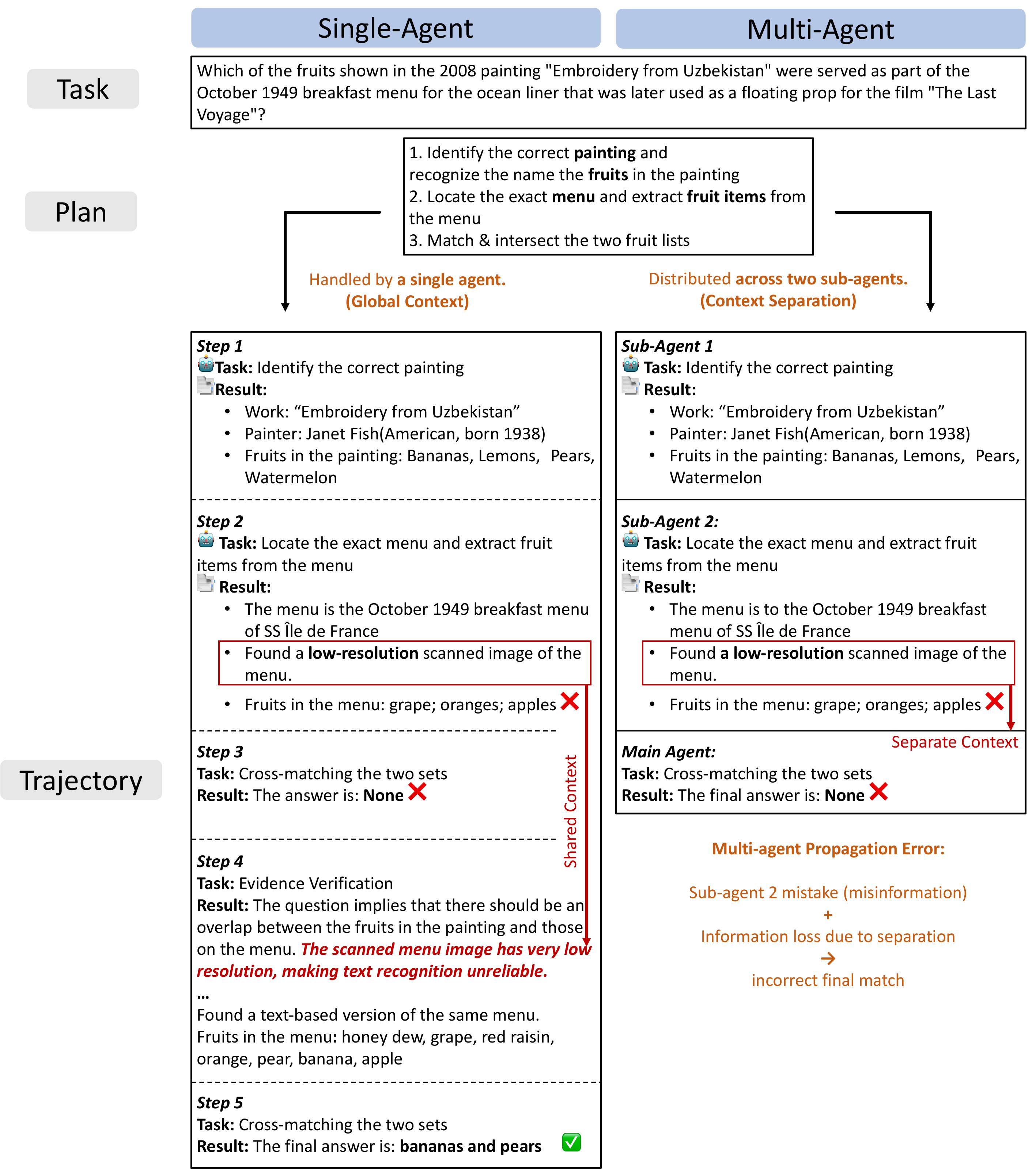}
\caption{\textbf{Mistake propagation in multi-agent.} Multi-agent structure has multiple different and separate context. Decomposing the sequential tasks has the risk of information loss or mistake propagation across agents, whereas a single-agent model has a more stable global context.
The red bounding box highlights where the agent made an error (caused by using a low-resolution image that produced a biased result). Because a single agent maintains more comprehensive contextual awareness, it can further correct and verify this error in subsequent tasks. In contrast, the downstream agent in a multi-agent setting may lack access to such fine-grained details, leading it to accept the incorrect conclusion.  \label{fig:gaia_single_vs_multi}}
\end{figure*}

\section{Agent Graph}
\label{sec:appendix_agent_graph}
\begin{figure*}[h]
\centering
\includegraphics[width=0.9\linewidth]{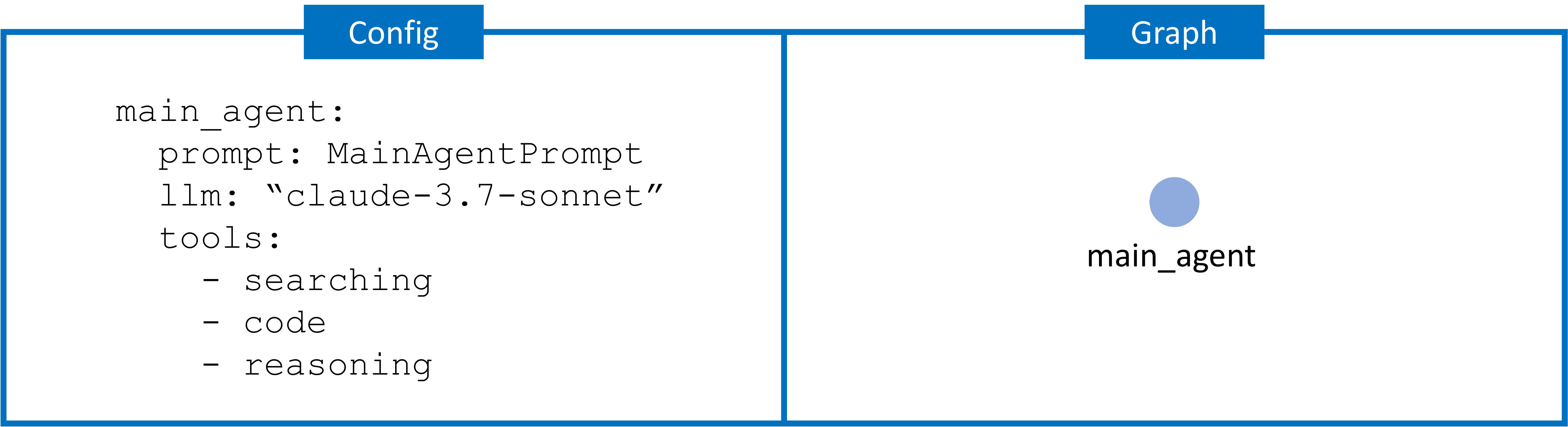}
\caption{\textbf{Single-Agent Baseline.} The system includes only one Main Agent, which must repeatedly perform reasoning steps, trigger tools, and call itself to complete a complex task. Due to the lack of role decomposition or specialized subagents, this architecture exhibits minimal flexibility and struggles with tasks that require modular expertise or parallel processing.
}
\label{fig:single_agent_graph}
\end{figure*}

In traditional agent architectures, two common forms are typically observed: single-agent and multi-agent systems. Specifically, as shown in Figure~\ref{fig:single_agent_graph}, a single-agent architecture consists of a single node where the task is handled by the agent itself. In this architecture, the agent repeatedly calls its own strategies and functions to gradually solve the problem. While this approach is simple and well-suited for clear, well-defined tasks, it is limited when dealing with more complex or dynamic scenarios.

\begin{figure*}[h]
\centering
\includegraphics[width=0.9\linewidth]{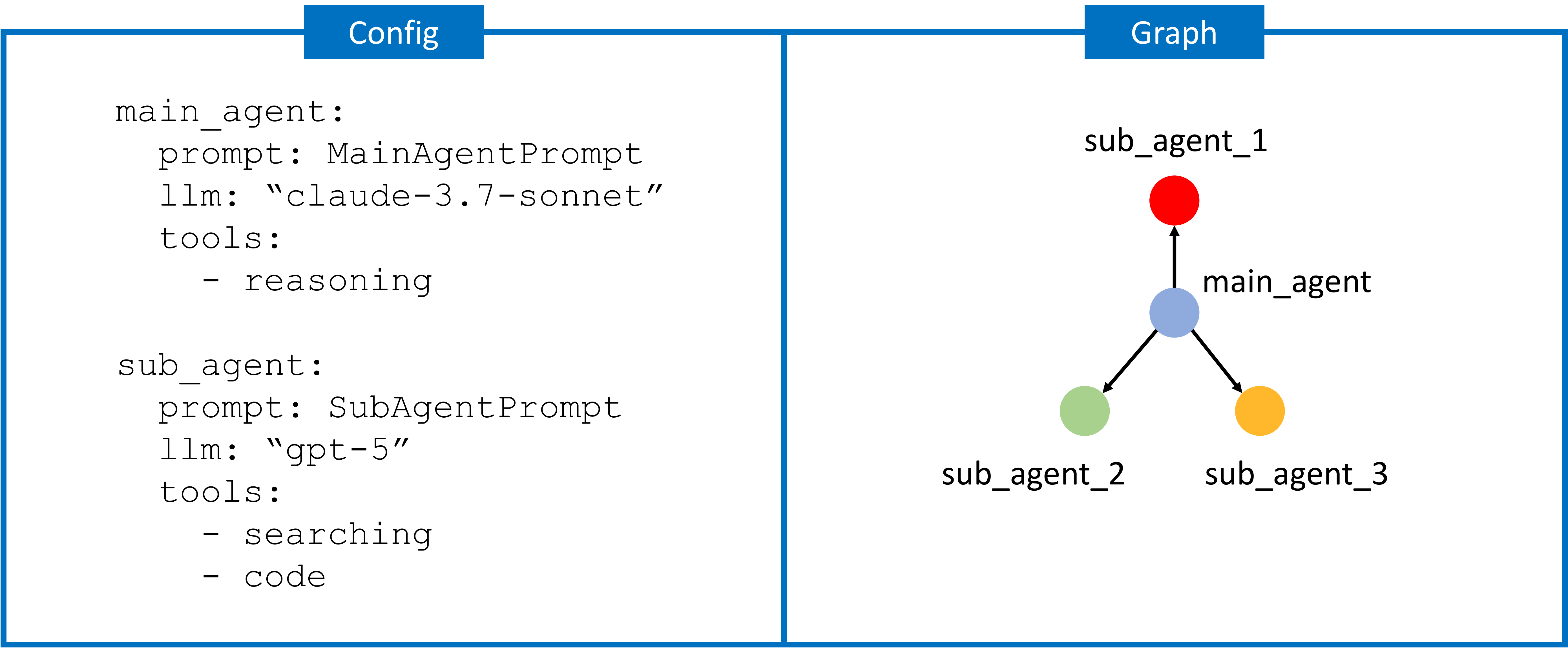}
\caption{\textbf{Multi-Agent Baseline.}The system consists of a Main Agent that decomposes a complex task into several subtasks and delegates them to multiple specialized Sub-Agents. Each Sub-Agent focuses on a specific type of reasoning or operation, enabling modularity, improved flexibility, and higher execution efficiency. Compared with the single-agent setup, this architecture better supports specialization, parallel processing, and robust task handling. However, it is still not flexible enough.}
\label{fig:multi_agent_graph}
\end{figure*}

On the other hand, a multi-agent system is more flexible and powerful, as shown in Figure~\ref{fig:multi_agent_graph}. In this setup, there is usually a main agent and multiple sub-agents. The main agent is responsible for overall task planning and scheduling, and it can repeatedly call different sub-agents to handle various parts of the task. Each sub-agent typically focuses on specific functions or tasks, allowing the system to achieve more efficient task decomposition and parallel processing. However, while this structure enhances flexibility and efficiency, it still faces challenges in terms of complexity and scalability, especially when the task size increases. Efficiently coordinating multiple sub-agents remains a key challenge.

In contrast to the traditional single-agent and multi-agent systems, our agent framework offers a more flexible and diverse call mechanism. We employ a directed graph structure that enables agents to interact and call upon each other in a more intricate manner. In this framework, task handling is not reliant on a single agent or a simple main-subordinate structure. Instead, it leverages a directed graph where a network of agents can dynamically interact with one another based on the task requirements. Each node in the graph represents an independent agent, and the relationships between agents are defined by edges, allowing agents to call upon each other’s functionalities and collaborate in a highly adaptable way.

The key advantage of this directed graph structure lies in its flexibility and scalability. Each agent can not only call upon its own strategies and functions but can also access the functions of other agents in the graph. This enables highly customized operations depending on the task at hand. For instance, when facing complex, multi-layered tasks, the main agent (or scheduling agent) can dynamically select appropriate sub-agents for task decomposition, while agents can exchange information and collaborate through the edges of the graph. This structure allows the system to efficiently handle complex tasks and adapt to changing environments.

\begin{figure*}[h]
\centering
\includegraphics[width=0.9\linewidth]{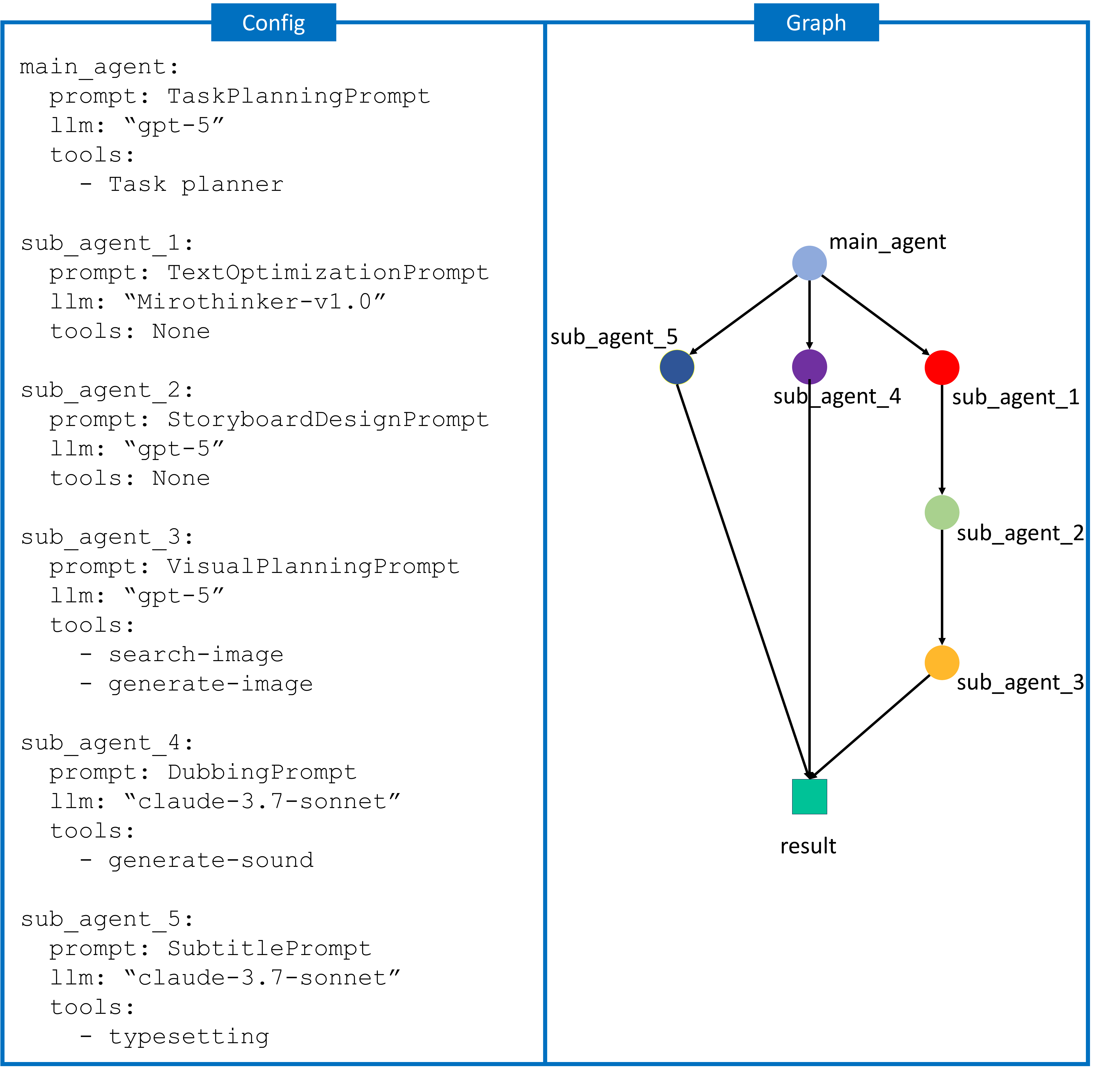}
\caption{\textbf{Agent Graph in short video generation.} The goal of the system is to transform a user-provided script into a complete short-video asset package. The Main Agent plans the workflow and dispatches subtasks to specialized Sub-Agents. Sub-Agent 1 optimizes and refines the input text, Sub-Agent 2 generates a structured storyboard, Sub-Agent 3 retrieves or creates visual materials, Sub-Agent 4 produces dubbing audio, and Sub-Agent 5 generates subtitles and handles typesetting. The outputs from all agents are finally aggregated to produce the final video content.
}
\label{fig:ShortVideoFeneration}
\end{figure*}

\begin{figure*}[h]
\centering
\includegraphics[width=0.9\linewidth]{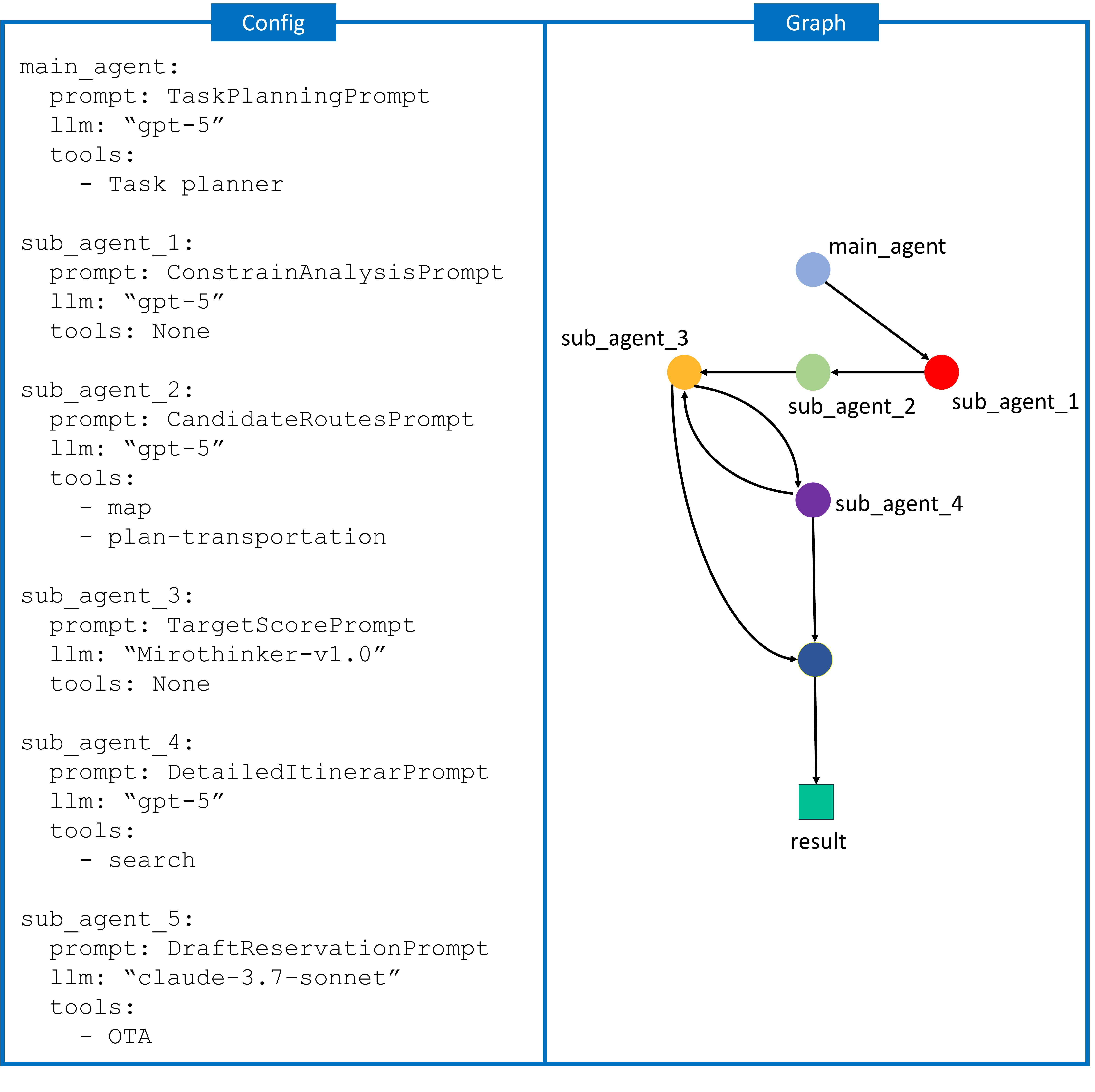}
\caption{\textbf{Agent Graph in complex travel planning and reservation drafting.} The goal of the system is to convert user constraints and travel preferences into a detailed multi-city itinerary and a reservation-ready plan. The Main Agent decomposes the task and coordinates specialized Sub-Agents. Sub-Agent 1 analyzes constraints such as time, budget, and travel rules; Sub-Agent 2 generates feasible candidate routes and transportation options; Sub-Agent 3 evaluates routes with scoring criteria; Sub-Agent 4 expands the selected plan into a detailed day-by-day itinerary; and Sub-Agent 5 drafts preliminary reservation information including flights, hotels, and activities. These outputs are merged to form the final travel plan delivered to the user.}
\label{fig:OTA}
\end{figure*}

\begin{figure*}[h]
\centering
\includegraphics[width=0.9\linewidth]{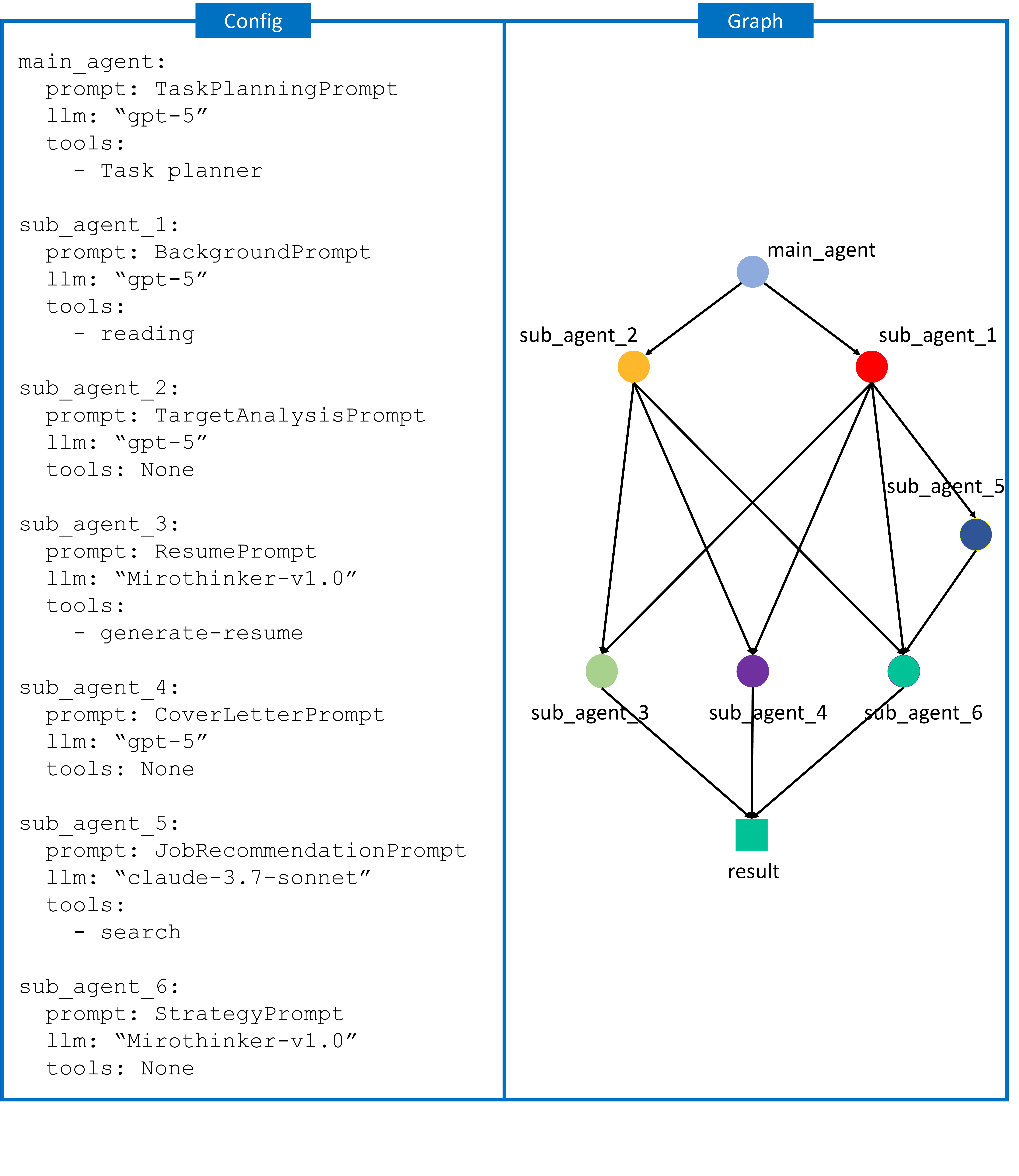}
\caption{\textbf{Agent Graph in job application material generation and delivery strategy.} The goal of the system is to transform a user’s background and target job preferences into a complete set of application materials and a personalized job-hunting strategy. The Main Agent plans the overall workflow and delegates subtasks to specialized Sub-Agents. Sub-Agent 1 analyzes the user’s background and extracts key experience, Sub-Agent 2 interprets job requirements and target role expectations, Sub-Agent 3 generates tailored resumes, Sub-Agent 4 produces customized cover letters, Sub-Agent 5 retrieves suitable job openings and recommendations, and Sub-Agent 6 formulates an optimized application strategy. All outputs are finally aggregated to produce the final job application package.}
\label{fig:Job}
\end{figure*}

Moreover, due to the high scalability of this architecture, when the system needs to tackle new tasks or extend its functionalities, new nodes can be added or the graph structure can be adjusted without the need to redesign the entire framework. This makes our agent framework more versatile and adaptable for a wide range of applications, with strong potential for long-term development. Three examples are shown in Figure~\ref{fig:ShortVideoFeneration}, ~\ref{fig:OTA}, ~\ref{fig:Job}
. It is important to note that Deep Research has a clearly linear, one-directional workflow structure; therefore, our method aligns with the multi-agent paradigm.

\section{Ablation on Context Length}
\paragraph{Context Length.} We investigate how maximum context length affects MiroFlow’s performance. Experiments are conducted on the GAIA validation set using GPT-5 in a single-agent configuration, which depends heavily on long contexts to maintain coherence across subtasks. We evaluate several context-length settings ranging up to 400k tokens, corresponding to GPT-5’s maximum input capacity.

\begin{figure}[h]
\centering
\includegraphics[width=0.6\linewidth]{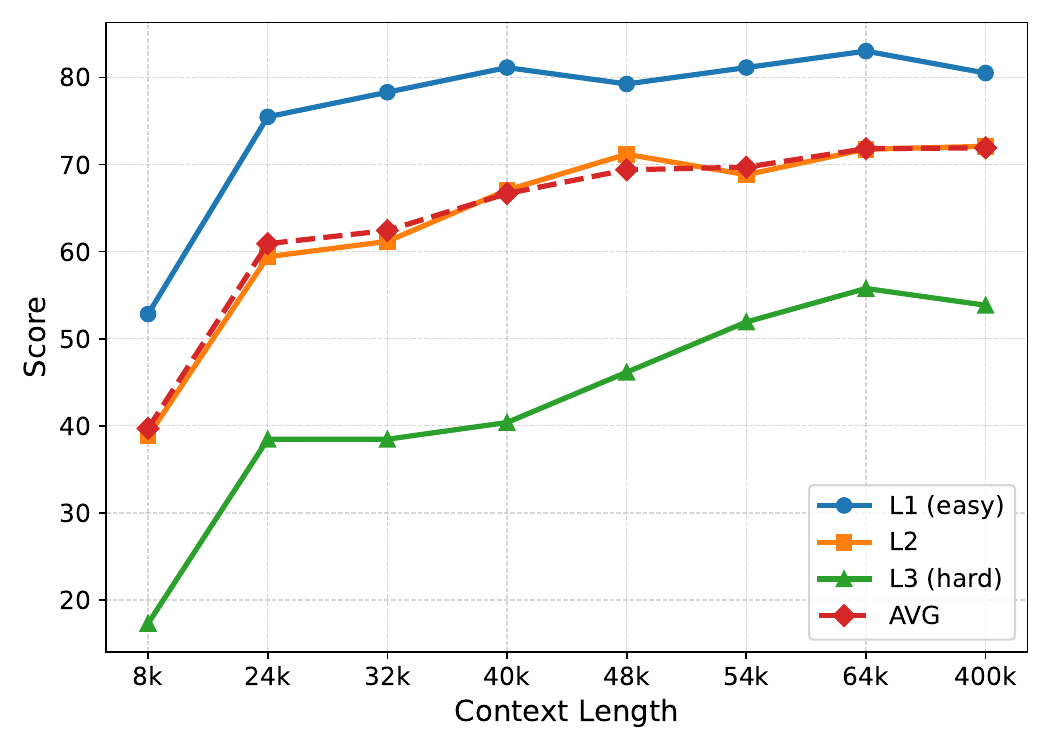}
\caption{\textbf{Agent performance with respect to LLM context length.} Performance of MiroFlow on the GAIA validation set using GPT-5 in a single-agent configuration under different maximum-context settings (8k–400k tokens). Accuracy improves substantially when increasing context from 8k to the 24k–48k range, after which gains begin to saturate. Harder tasks (L3) exhibit the strongest dependence on longer contexts.}
\label{fig:context_ablation}
\end{figure}

As shown in Figure~\ref{fig:context_ablation}, performance improves steadily as the context length increases. Short contexts (e.g., 8k) limit the model’s ability to retain intermediate reasoning steps, leading to incomplete or inconsistent solutions. Once the context is sufficiently large (24k–48k), accuracy rises notably across all difficulty levels. Beyond this range, gains gradually saturate, with only marginal improvements when extending to very long contexts such as 400k tokens. Harder tasks (L3) benefit the most from additional context, reflecting their greater dependence on long-range reasoning.

\section{Agent Prompts}

System prompt for the main agent in benchmark experiments:

\begin{promptbox}
\small\ttfamily
In this environment you have access to a set of tools you can use to answer the user's question. 
\newline \\
You only have access to the tools provided below. You can only use one tool per message, and will receive the result of that tool in the user's next response. You use tools step-by-step to accomplish a given task, with each tool-use informed by the result of the previous tool-use. Today is: \{formatted\_date\}
\newline \\
\# Tool-Use Formatting Instructions 
\newline \\
Tool-use is formatted using XML-style tags. The tool-use is enclosed in <use\_mcp\_tool></use\_mcp\_tool> and each parameter is similarly enclosed within its own set of tags.
\newline \\
The Model Context Protocol (MCP) connects to servers that provide additional tools and resources to extend your capabilities. You can use the server's tools via the `use\_mcp\_tool`.
\newline \\
Description: 
Request to use a tool provided by a MCP server. Each MCP server can provide multiple tools with different capabilities. Tools have defined input schemas that specify required and optional parameters.
\newline \\
Parameters:\\
- server\_name: (required) The name of the MCP server providing the tool\\
- tool\_name: (required) The name of the tool to execute\\
- arguments: (required) A JSON object containing the tool's input parameters, following the tool's input schema, quotes within string must be properly escaped, ensure it's valid JSON
\newline \\
Usage:\\
<use\_mcp\_tool>\\
<server\_name>server name here</server\_name>\\
<tool\_name>tool name here</tool\_name>\\
<arguments>\\
\{\{
"param1": "value1",
"param2": "value2 \textbackslash"escaped string\textbackslash""
\}\}
</arguments>\\
</use\_mcp\_tool>
\newline \\
Important Notes:\\
- Tool-use must be placed **at the end** of your response, **top-level**, and not nested within other tags.\\
- Always adhere to this format for the tool use to ensure proper parsing and execution.
\newline \\
String and scalar parameters should be specified as is, while lists and objects should use JSON format. Note that spaces for string values are not stripped. The output is not expected to be valid XML and is parsed with regular expressions.\\
Here are the functions available in JSONSchema format:
\newline \\
\# General Objective
\newline \\
You accomplish a given task iteratively, breaking it down into clear steps and working through them methodically.
\newline \\
\#\# Task Strategy
\newline \\
1. Analyze the user's request and set clear, achievable sub-goals. Prioritize these sub-goals in a logical order.\\
2. Start with a concise, numbered, step-by-step plan (e.g., 1., 2., 3.) outlining how you will solve the task before taking any action. Each sub-goal should correspond to a distinct step in your task-solving process.\\
3. Work through these sub-goals sequentially. After each step, carefully review and extract all potentially relevant information, details, or implications from the tool result before proceeding. The user may provide tool-use feedback, reflect on the results, and revise your plan if needed. If you encounter new information or challenges, adjust your approach accordingly. Revisit previous steps to ensure earlier sub-goals or clues have not been overlooked or missed.\\
4. You have access to a wide range of powerful tools. Use them strategically to accomplish each sub-goal.
\newline \\
\#\# Tool-Use Guidelines
\newline \\
1. **IMPORTANT: Each step must involve exactly ONE tool call only, unless the task is already solved. You are strictly prohibited from making multiple tool calls in a single response.** \\
2. Before each tool call:\\
- Briefly summarize and analyze what is currently known.\\
- Identify what is missing, uncertain, or unreliable.\\
- Be concise; do not repeat the same analysis across steps.\\
- Choose the most relevant tool for the current sub-goal, and explain why this tool is necessary at this point.\\
- Verify whether all required parameters are either explicitly provided or can be clearly and reasonably inferred from context.\\
- Do not guess or use placeholder values for missing inputs.\\
- Skip optional parameters unless they are explicitly specified.\\
3. All tool queries must include full, self-contained context. Tools do not retain memory between calls. Include all relevant information from earlier steps in each query.\\
4. Avoid broad, vague, or speculative queries. Every tool call should aim to retrieve new, actionable information that clearly advances the task.\\
5. **For historical or time-specific content**: Regular search engines return current webpage content, not historical content. Archived webpage search is essential for retrieving content as it appeared in the past, use related tools to search for the historical content.\\
6. Even if a tool result does not directly answer the question, thoroughly extract and summarize all partial information, important details, patterns, constraints, or keywords that may help guide future steps. Never proceed to the next step without first ensuring that all significant insights from the current result have been fully considered.
\newline \\
\#\# Tool-Use Communication Rules
\newline \\
1. **CRITICAL: After issuing exactly ONE tool call, STOP your response immediately. You must never make multiple tool calls in a single response. Do not include tool results, do not assume what the results will be, and do not continue with additional analysis or tool calls. The user will provide the actual tool results in their next message.**\\
2. Do not present the final answer until the entire task is complete.\\
3. Do not mention tool names.\\
4. Do not engage in unnecessary back-and-forth or end with vague offers of help. Do not end your responses with questions or generic prompts.\\
5. Do not use tools that do not exist.\\
6. Unless otherwise requested, respond in the same language as the user's message.\\
7. If the task does not require tool use, answer the user directly.\\
\end{promptbox}

\end{document}